\documentclass{article}

\usepackage{arxiv}

\usepackage[utf8]{inputenc} 
\usepackage[T1]{fontenc}    
\usepackage{url}            
\usepackage{booktabs}       
\usepackage{amsfonts}       
\usepackage{nicefrac}       
\usepackage{microtype}      
\usepackage{lipsum}
\usepackage{amsmath}
\usepackage{amsthm}
\usepackage{amssymb}   
\usepackage{bigints}
\usepackage{bm}
\usepackage{diagbox}
\usepackage{makecell}
\usepackage{multirow}
\usepackage{caption}
\usepackage{subcaption}
\usepackage{graphicx}
\usepackage[ruled,vlined]{algorithm2e}
\include{pythonlisting}
\usepackage{xcolor}
\usepackage{mathtools}
\usepackage{enumitem}

\usepackage{hyperref}
\hypersetup{
    colorlinks=true,
    citecolor=black,
    linkcolor=black,
    filecolor=black,
    urlcolor=black,
}

\newcommand{\R}{\mathbb{R}}

\theoremstyle{definition}

\numberwithin{equation}{section}

\theoremstyle{plain}

\title{Fast PDE-constrained optimization via self-supervised operator learning}

\author{
  Sifan Wang \\
  Graduate Group in Applied Mathematics \\
  and Computational Science \\
  University of Pennsylvania\\
  Philadelphia, PA 19104 \\
  \texttt{sifanw@sas.upenn.edu} \\
   \And
     Mohamed Aziz Bhouri \\
  Department of Mechanichal Engineering \\
  and Applied Mechanics\\
  University of Pennsylvania\\
  Philadelphia, PA 19104 \\
  \texttt{bhouri@seas.upenn.edu} \\
  \And
  Paris Perdikaris \\
  Department of Mechanichal Engineering \\
  and Applied Mechanics\\
  University of Pennsylvania\\
  Philadelphia, PA 19104 \\
  \texttt{pgp@seas.upenn.edu} \\
}

\hypersetup{
pdftitle={A template for the arxiv style},
pdfsubject={q-bio.NC, q-bio.QM},
pdfauthor={David S.~Hippocampus, Elias D.~Striatum},
pdfkeywords={First keyword, Second keyword, More},
}

\begin{document}
\maketitle

\begin{abstract}
Design and optimal control problems are among the fundamental, ubiquitous tasks we face in science and engineering. In both cases, we aim to represent and optimize an unknown (black-box) function that associates a performance/outcome to a set of controllable variables through an experiment. In cases where the experimental dynamics can be described by partial differential equations (PDEs), such problems can be mathematically translated into PDE-constrained optimization tasks, which quickly become intractable as the number of control variables and the cost of experiments increases. In this work we leverage physics-informed deep operator networks (DeepONets) -- a self-supervised framework for learning the solution operator of parametric PDEs -- to build fast and differentiable surrogates for rapidly solving PDE-constrained optimization problems, even in the absence of any paired input-output training data. The effectiveness of the proposed framework will be demonstrated across different applications involving continuous functions as control or design variables, including time-dependent optimal control of heat transfer, and drag minimization of obstacles in Stokes flow. In all cases, we observe that DeepONets can minimize high-dimensional cost functionals in a matter of seconds, yielding a significant speed up compared to traditional adjoint PDE solvers that are typically costly and limited to relatively low-dimensional control/design parametrizations. 
\end{abstract}


\section{Introduction}
 
{\bf Motivation:}
A ubiquitous task across many disciplines in science and engineering is to optimize the performance of a system governed by physical laws that are mathematically expressed as systems of partial differential equations (PDEs). Many of such tasks can be formally formulated as  PDE-constrained optimization problems which naturally arise in various settings such as shape optimization \cite{sokolowski1992introduction,haslinger2003introduction}, optimal control \cite{lewis2012optimal}, image processing \cite{de2013image}, aerodynamics \cite{hicks1978wing,frank1992comparison}, crystal growth \cite{ng2012optimal}, drug delivery \cite{chakrabarty2005optimal}, heat transfer phenomena  \cite{fabbri2000heat,chen2009optimization} and finance \cite{cornuejols2006optimization}. Despite their prominence, PDE-constrained optimization problems are notoriously hard to solve, especially in cases where the cost of simulating the PDE system is high and the number of design/control parameters is large. The latter can quickly lead to intractable complexity, often referred to as the {\em curse of dimensionality}, as a naive exploration of the decision space via grid search requires a number of experiments that grow exponentially with the dimension of the design/control parameters. Even more severe is the setting where those design/control parameters are not finite-dimensional vectors, but continuous functions with infinitely many degrees of freedom. 

{\bf Related work:}
A large volume of literature has focused on designing effective solutions for tackling PDE-constrained optimization problems, primarily by developing fast emulators and surrogate models that can bypass the cost of expensive experiments \cite{audet2000surrogate,queipo2005surrogate,forrester2009recent}, as well as formulating effective strategies for exploring the input parameter space, for example using gradient-based methods \cite{kennedy2014parallel}, adjoint solvers \cite{Beatson2020neural,Tianju2020FEPDE}, or Bayesian optimization \cite{ryan2016review,lam2018advances}. Fast emulators are typically built by reduced-order modeling techniques \cite{lucia2004reduced,quarteroni2014reduced} formulated on the premise that solutions of parametric PDEs live on a low-dimensional manifold. Given a (typically large) number of high-fidelity realizations of the original system that sufficiently span the parameter space, one can discover the low-dimensional manifold using techniques such as principal components analysis \cite{sirovich1987turbulence,willcox2002balanced,audouze2009reduced}, diffusion maps \cite{coifman2008diffusion}, or auto-encoders \cite{lee2020model,maulik2021reduced,xu2020multi}.  
If successful, then one can simulate the system's performance given a new design/control input using far less degrees of freedom compared to the full-order experiment, yielding often dramatic computational speed-ups, but often at the price of reduced fidelity \cite{Herzog2010alg,amsallem2008interpolation}.
While reduced-order models are currently one of the main workhorses for  PDE-constrained optimization \cite{lucia2004reduced,biegler2007real}, their construction can be computationally expensive as it requires accumulating a large number of system responses to input excitations. Furthermore, they usually lack robustness with respect to parameter changes and therefore must often be rebuilt for each parameter variation \cite{amsallem2008interpolation}. As such, their effectiveness can quickly deteriorate as the number of input design/control variables increases.

More recently, deep learning approaches have been used to tackle PDE-constrained tasks across diverse applications, including electromagnetism \cite{sasaki2019topology}, nanophotonics \cite{so2020deep}, material science \cite{guo2021artificial},  structural mechanics \cite{hoyer2019neural}, and optimal control \cite{bansal2021deepreach}. However, many of these approaches still rely on traditional numerical solvers and require an excessive amount of training data that can be costly to obtain, while their predictions are typically not consistent with the underlying PDE that generated the training data \cite{wang2021learning}. The latter can be addressed by adopting frameworks such as physics-informed neural networks (PINNs) \cite{raissi2019deep, lu2021physics, hennigh2021nvidia}. However, PINNs are known to be notoriously hard to train \cite{wang2021understanding, wang2020and,wang2021eigenvector}, and can incur a large computational cost during the online phase of a PDE-constrained optimization task as they may need to be re-trained in order to accurately generalize across different design/control inputs.

{\bf Contributions of this work:}
In this work we propose to leverage physics-informed deep operator networks (DeepONets) \cite{wang2021learning,wang2021long,wang2021improved}; a recently developed deep learning framework for learning the solution operator of parametric PDEs in a self-supervised manner. This enables the construction of effective emulators for querying the solution of PDEs at an arbitrary spatio-temporal resolution, given continuous (i.e. infinite-dimensional) design/control inputs, and without the requirement of generating large training data-sets via repeatedly evaluating costly conventional simulators. As we shall see, a trained  physics-informed DeepONet model can be used as a fast and differentiable surrogate for tackling high-dimensional PDE-constrained optimization problems via gradient-based optimization in near real-time. The effectiveness of the proposed framework will be demonstrated through different case studies in optimal control and shape optimization, demonstrating a significant speed up compared to traditional adjoint finite element solvers.

{\bf Structure of this paper:}
In section \ref{sec: methods}, we formulate general PDE-constrained optimization tasks and present a brief overview of physics-informed DeepONets. Next, we describe how a trained physics-informed DeepONet model can be leveraged as a fast and differentiable surrogate for rapidly  solving PDE-constrained optimization problems. In section \ref{sec: results}, we perform a series numerical studies to demonstrate the effectiveness of the proposed framework. Finally, section \ref{sec: summary}  concludes with a discussion of our main findings, potential pitfalls, and shortcomings, as well as future research directions emerging from this study. All code and data accompanying this manuscript will be made available at \url{https://github.com/PredictiveIntelligenceLab/PDE-constrained-optimization-PI-DeepONet}.

\section{Methods}
\label{sec: methods}

\subsection{PDE-constrained optimization}
\label{sec: setup}
We begin with defining PDE-constrained optimization problems in an abstract form. Let $U, S, V$ be Banach spaces, and consider the following general PDE-constrained optimization problem
\begin{align}
    \label{eq: opt_cost_func}
    \min \quad &\mathcal{J}(\bm{u}, \bm{s})\\
    \label{eq: opt_pde_constrain}
    \textrm{s.t.} \quad & \mathcal{E}(\bm{u}, \bm{s}) = 0 \\
    \label{eq: opt_constrain}
    & \bm{u} \in U_{ad}, \bm{s} \in S_{ad},
\end{align}
where $\mathcal{J}: U \times S \rightarrow \mathbb{R}$ denotes a cost function to be minimized, and  $\mathcal{E} : U \times S \rightarrow V$ represents a system of PDEs subject to appropriate initial and boundary conditions. Moreover, the latent state $\bm{s}$ that satisfies the PDE system is assumed to belong to an admissible space of functions $S_{ad} \subset S$, while the associated control/input functions $\bm{u}$ belong to an admissible function space $U_{ad} \subset U$. Here we assume that for every $\bm{s} \in S_{ad}$, there exists a unique $\bm{u} = \bm{u}(\bm{s})$ such that $\mathcal{E}(\bm{u}, \bm{s}) = 0$. Consequently, we can  define the PDE solution operator $G: U \rightarrow S$ that maps input control functions $\bm{u}$ to their associated PDE solutions $\bm{s}$ as $G(\bm{u}) = \bm{s}$.

\subsection{Physics-informed DeepONets}

Deep operator networks (DeepONets) \cite{lu2021learning} provide a specialized deep learning framework that aims to learn abstract nonlinear operators between infinite-dimensional function spaces. The architecture is inspired and validated by the rigorous universal approximation theorem for operators \cite{chen1995universal, lu2021learning}. In follow up work, Wang {\it et. al.} \cite{wang2021learning} developed  physics-informed DeepONets, introducing a simple and effective regularization mechanism for biasing the outputs of DeepONet models towards ensuring physical consistency. Here we give a brief overview of physics-informed DeepONets, with a special focus on learning the solution operator of parametric PDEs. The terminology "parametric PDEs" refers to the fact that some parameters of a given PDE system are allowed to vary over a certain range. These input parameters may include, but are not limited to, the shape of the physical domain, the initial or boundary conditions, constant or variable coefficients (e.g. diffusion or reaction rates), source terms, etc. To describe such problems in their full generality, let $(\mathcal{U}, \mathcal{V}, \mathcal{S})$ be a triplet of Banach spaces, and $\mathcal{N}: \mathcal{U} \times \mathcal{S} \rightarrow \mathcal{V}$ be a linear or nonlinear differential operator. We consider general parametric PDEs taking the form
\begin{align}
    \label{eq: parametric_PDE}
    \mathcal{N}(\bm{u}, \bm{s})(\bm{x}) = 0,
\end{align}
subject to boundary conditions
\begin{align}
    \mathcal{B}(\bm{u}, \bm{s})(\bm{x}) = 0,
\end{align}
where $\bm{u} \in \mathcal{U}$ denotes the parameters (i.e. input functions), and $\bm{s} \in \mathcal{S}$ denotes the unknown latent quantity of interest that is governed by the  PDE system of equation \ref{eq: parametric_PDE}. Moreover, $\mathcal{N}[\cdot]$ is a differential operator, and  $\mathcal{B}[\cdot] $ denotes a boundary conditions operator that enforces any Dirichlet, Neumann, Robin, or periodic boundary conditions. For time-dependent problems, we consider time $t$ as a special component of $\bm{x}$, and $\Omega$ is extended to contain the temporal domain. In that case, initial conditions can be simply treated as a special type of boundary condition defined in the joint spatio-temporal domain.

Having assumed that, for any $\bm{u} \in \mathcal{U}$, there exists an unique solution $\bm{s} = \bm{s}(\bm{u}) \in \mathcal{U}$ to \ref{eq: parametric_PDE} (subject to appropriate initial and boundary conditions), then we can define the solution operator $G: \mathcal{U} \rightarrow \mathcal{S}$ as
\begin{align}
    G(\bm{u}) = \bm{s}(\bm{u}).
\end{align}
Following the original formulation of Lu {\it et al.} \cite{lu2021learning},  the solution map $G$ can be represented by an unstacked DeepONet $G_{\bm{\theta}}$, where $\bm{\theta}$ denotes all trainable parameters (i.e. weights and biases) of the DeepONet network. 
As shown in Figure \ref{fig: arch}, the DeepONet architecture is composed of two separate neural networks referred to as the ``branch" and ``trunk" networks, respectively. The branch network takes $\bm{u}$ as inputs and returns a features embedding $[b_1, b_2,\dots, b_q]^T \in \mathbb{R}^q$ as output, where $\bm{u} = [\bm{u}(\bm{x}_1), \bm{u}(\bm{x}_2), \dots, \bm{u}(\bm{x}_m) ]$ represents a function $\bm{u} \in \mathcal{U}$ evaluated at a collection of fixed locations $\{\bm{x}_i\}_{i=1}^m$.  The trunk network takes the continuous coordinates $\bm{y}$ as inputs, and constructs a features embedding $[t_1, t_2,\dots, t_q]^T \in \mathbb{R}^q$. The final output is obtained by performing the dot product of the outputs of the branch and trunk networks. Specifically, a DeepONet $G_{\bm \theta}$ prediction of a function $\bm{u}$ evaluated at $\bm{y}$  can be expressed by
\begin{align}
    G_{\bm{\theta}}(\bm{u})(\bm{y}) = \sum_{k=1}^{q} \underbrace{b_{k}\left(\bm{u}\left(\bm{x}_{1}\right), \bm{u}\left(\bm{x}_{2}\right), \ldots, \bm{u}\left(\bm{x}_{m}\right)\right)}_{\text {branch }} \underbrace{t_{k}(\bm{y})}_{\text {trunk }}.
\end{align}
Then, a physics-informed DeepONet model can be trained by minimizing the following composite loss function
\begin{align}
    \label{eq: physics_informed_loss}
    \mathcal{L}(\bm{\theta}) = \mathcal{L}_{\text{BC}}(\bm{\theta}) +  \mathcal{L}_{\text{PDE}}(\bm{\theta}),
\end{align}
where 
\begin{align}
    \label{eq: loss_bc}
    &\mathcal{L}_{\text{BC}}(\bm{\theta}) = \frac{1}{NP} \sum_{i=1}^N \sum_{j=1}^P \left|\mathcal{B}(u^{(i)}, G_{\bm{\theta}} (\bm{u}^{(i)}) )(\bm{y}^{(i)}_{b,j}) \right|^2  \\
    \label{eq: loss_r}
    &  \mathcal{L}_{\text{PDE}}(\bm{\theta}) = \frac{1}{NQ} \sum_{i=1}^N \sum_{j=1}^Q  \left|\mathcal{N}(u^{(i)}, G_{\bm{\theta}}(\bm{u}^{(i)})(\bm{y}^{(i)}_{r,j}) ) \right|^2.
\end{align}
Here,  $\{ \bm{u}^{(i)} \}_{i=1}^N$ denotes $N$ separate input functions sampled from $\mathcal{U}$. For each  $\bm{u}^{(i)}$, $\{\bm{y}^{(i)}_{b,j}\}_{j=1}^P$ are $P$ locations that are determined by the boundary conditions. Besides,  $\{\bm{y}^{(i)}_{r,j}\}_{j=1}^Q$ is a set of  collocation points that can be randomly sampled in the domain of $G(\bm{u}^{(i)})$. 
As a consequence, $\mathcal{L}_{\text{BC}}(\bm{\theta})$ fits the initial/boundary conditions, while  $\mathcal{L}_{\text{PDE}}(\bm{\theta})$ enforces the underlying PDE constraints. It is worth emphasizing  that the formulation of the physics-informed DeepONets does not require any paired input-output observations for training the model, except for a set of given boundary conditions \cite{wang2021learning}. We also remark that the DeppONet architecture yields a continuous representation of the output functions that can be differentiated with respect to their input coordinates, thus allowing  us to directly compute any given differential constraints via automatic differentiation \cite{griewank1989automatic, baydin2018automatic}.

\subsection{PDE-constrained optimization with physics-informed DeepONets}
In this section, we describe our proposed method for solving general  PDE-constrained optimization problems (see equations \ref{eq: opt_cost_func} - \ref{eq: opt_constrain}) using physics-informed DeepONets. As summarized in Algorithm \ref{alg: alg}, the proposed framework consists of two main steps. 

{\bf Step 1:} In the first step, we represent the PDE solution operator $G: U \rightarrow S$ by a physics-informed DeepONet $G_\theta$, which is trained  by minimizing the composite loss function in equation \ref{eq: physics_informed_loss}. For every input control function $\bm{u} \in U_{ad}$, a trained physics-informed DeepONet can rapidly return the corresponding PDE solution $s(\bm{u})$ via a simple model evaluation. 
Moreover, the DeepONet model predictions are differentiable with respect to the input control functions, allowing us to employ gradient-based optimization to expediently minimize the cost functional $\mathcal{J}$ in equation \ref{eq: opt_cost_func}.

\begin{figure}
    \centering
    \includegraphics[width=0.9\textwidth]{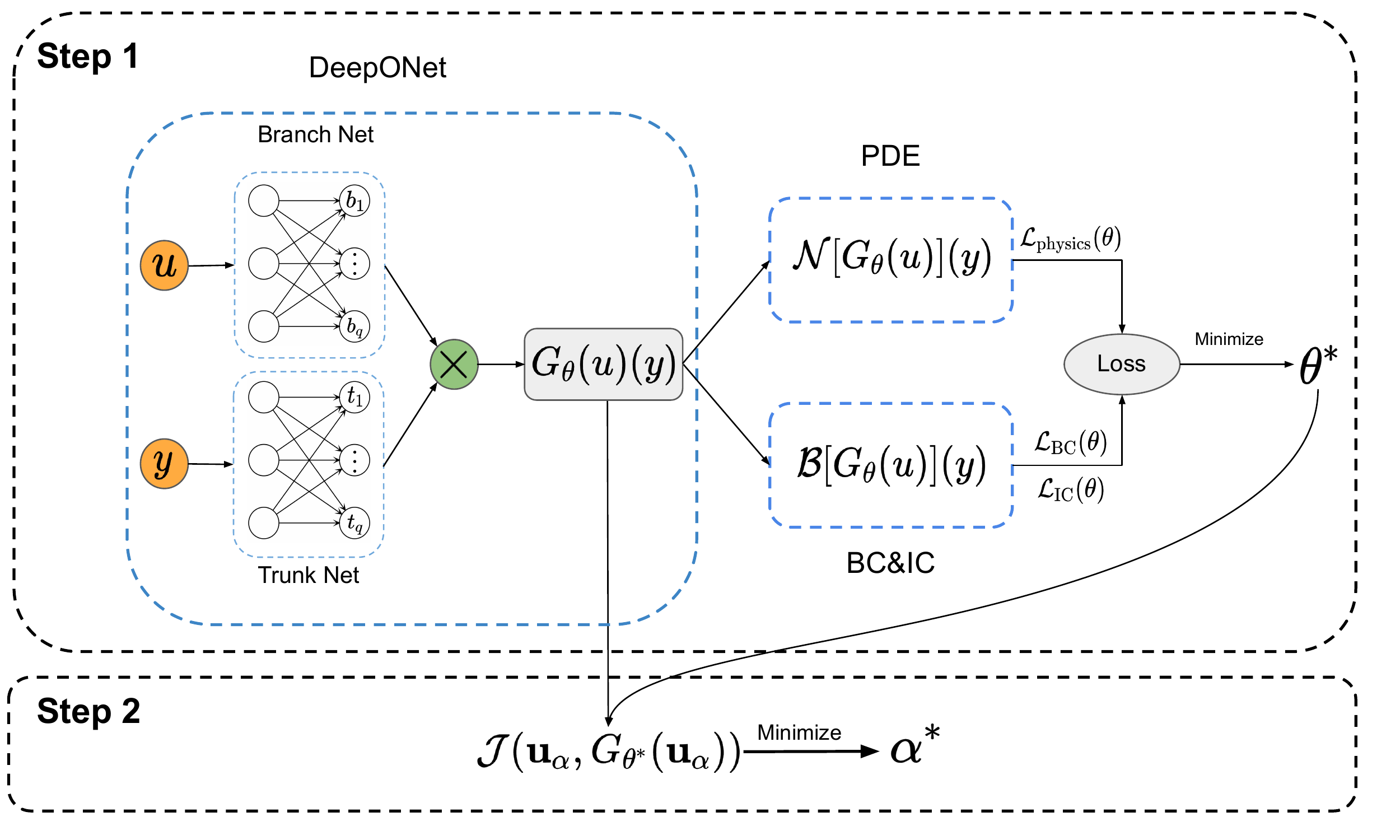}
    \caption{{\em PDE-constrained optimization workflow:} In Step 1, a physics-informed DeepONet $G_{\bm{\theta}}$ is trained to learn the solution operator for the given parametric PDE system \ref{eq: opt_pde_constrain}. In Step 2, a neural network $u_{\bm{\alpha}}$ is employed to represent the control function and then trained by minimizing the cost function $\mathcal{J}$ defined in \ref{eq: opt_cost_func} where the trained physics-informed DeepONet is used as a surrogate forward PDE solver. It is worth noting that the algorithm does not require any solution measurements except for a set of boundary and initial conditions.  }
    \label{fig: arch}
\end{figure}

{\bf Step 2:} In the second step, we construct a parametrization of the input control functions $u_{\bm{\alpha}}$ (e.g. a deep neural network, or otherwise) with tunable parameters $\bm{\alpha}$, and minimize the cost function \ref{eq: opt_cost_func} using gradient descent to identify a locally optimal solution. An illustration of the proposed workflow is presented in Figure \ref{fig: arch}.

\begin{algorithm}
\SetAlgoLined
{\bf Step 1:} Train a physics-informed DeepONet to learn the PDE solution operator $G: U \rightarrow S$.
\begin{enumerate}[label=(\alph*), leftmargin=*]
   \item Represent the PDE solution operator by a DeepONet $G_{\bm{\theta}}$ with parameters $\bm{\theta}$.
   \item Prepare the training data for the PDE constraints and boundary conditions according to the system of governing equations  $\mathcal{E}(\bm{u}, \bm{s})$.
   \item Formulate a physics-informed loss function $\mathcal{L}(\bm{\theta})$ by considering both the PDE and the boundary condition residuals.
   \item Train the physics-informed DeepONet to identify a set of optimal  parameters $\bm{\theta}^*$ by minimizing the loss function $\mathcal{L}(\bm{\theta})$  via a stochastic gradient descent method.
\end{enumerate} 

{\bf Step 2:} Train a neural network to minimize the cost function $\mathcal{J}$. 
\begin{enumerate}[label=(\alph*), leftmargin=*]
    \item Parametrize the input control functions $\bm{u}_{\bm{\alpha}}$, where $\bm{\alpha}$ is a set of tunable parameters that need to be optimized (e.g. the weights and biases of a deep neural network).
    \item Formulate a loss function $\mathcal{J}(\bm{\alpha})$ using the trained physics-informed DeepONet $G_{\bm{\theta}^*}$ according to the definition of the cost functional in equation \ref{eq: opt_cost_func}.
    \item Minimizing the cost functional $\mathcal{J}(\bm{\alpha})$ via a gradient descent method to identify the optimal control parameters $\bm{\alpha}^{\ast}$.
\end{enumerate} 
$^{\ast}$All gradients (i.e. $\partial{\mathcal{L}}/\partial\bm{\theta}$,  $\partial{\mathcal{J}}/\partial\bm{\alpha}$ and $\partial{\bm{s}}/\partial\bm{x}$) are computed using automatic differentiation).

 \caption{PDE-constrained optimization using physics-informed DeepONets.}
 \label{alg: alg}
\end{algorithm}

\section{Results}
\label{sec: results}

In the following, we demonstrate the effectiveness of physics-informed DeepONets across a series of numerical studies involving various types of PDE-constrained optimization problems, including optimal control and shape optimization.
The hyper-parameter settings, performance metrics, and validation details are discussed in Appendix \ref{app: hp_settings},  \ref{app: metrics}, and \ref{app: validation}, respectively.  It is worth emphasizing that, in all cases, the proposed deep learning models are trained without any paired input-output data, assuming only knowledge of the underlying governing PDE system and its corresponding initial or boundary conditions.  All code and data accompanying this manuscript will be made publicly available at \url{https://github.com/PredictiveIntelligenceLab/PDE-constrained-optimization-PI-DeepONet}.

\subsection{Optimal control of a 1D Poisson equation}

We start with a pedagogical example involving optimal control of a 1D Poisson equation. Physically, this problem can the interpreted as finding the optimal heating/cooling of a cooktop to achieve the desired temperature profile. Mathematically, the objective is to minimize the following functional
\begin{align}
  \mathcal{J}(u, s) =  \quad & \frac{1}{2} \int_{\Omega}(s-d)^{2} \mathrm{~d} x+\frac{\alpha}{2} \int_{\Omega} u^{2} \mathrm{~d} x,
\end{align}
subject to
\begin{align}
\label{eq: poisson_eq}
- \kappa \Delta s = u \quad & \text { in } \Omega, \\
\label{eq: poisson_bc}
s =0  \quad  & \text { on } \partial \Omega,
\end{align}
where $\Omega$ is a bounded domain, $s: \Omega \rightarrow \R$ is the unknown temperature satisfying equation \ref{eq: poisson_eq}, subject to the boundary conditions prescribed in equation \ref{eq: poisson_bc}. Moreover,  $d: \Omega \rightarrow \R$ is the given desired temperature profile, while $f: \Omega \rightarrow \R$ is the unknown control function acting as source term. 
In addition, $\kappa \in \R$ is the thermal diffusivity, and $\alpha \geq 0$ is a Tikhonov regularization parameter, which is one of the most commonly used methods of regularization of ill-posed problems \cite{vogel2002computational}. In this example, we set $\Omega = [0, 1]$, $\kappa=1$, $\alpha = 0$, and $d(x) = \frac{1}{\pi^2} \sin(\pi x)$. Consequently, $u(x) = \sin(\pi x)$ is the global minimum at which we have $\mathcal{J} = 0$. 

Following the first step of Algorithm \ref{alg: alg}, we  use a DeepONet $G_{\bm{\theta}}$ to represent the solution operator $G$ mapping the source term $u$ to the associated PDE solution $s$, where the branch network and the trunk network are two separate 5-layer modified multi-layer perceptrons (MLPs) with 100 neurons per hidden layer and hyperbolic tangent activation functions (see Appendix \ref{app: modified_deeponet} for details). According to equations \ref{eq: poisson_eq} - \ref{eq: poisson_bc}, a physics-informed loss function can be formulated as follows
\begin{align}
    \mathcal{L}(\bm{\theta}) = \mathcal{L}_{\text{BC}}(\bm{\theta}) + \mathcal{L}_{\text{PDE}}(\bm{\theta}), 
\end{align}
where
\begin{align}
    &\mathcal{L}_{\text{BC}}(\bm{\theta}) = \frac{1}{NP} \sum_{i=1}^N  \sum_{j=1}^P \left| G_{\bm{\theta}}(\bm{u}^{(i)})(y_{b, j}^{(i)})   \right|^2, \\
    &\mathcal{L}_\text{PDE}(\bm{\theta}) = \frac{1}{NQ} \sum_{i=1}^N  \sum_{j=1}^Q \left| \frac{\partial^2 G_{\bm{\theta}}(\bm{u}^{(i)})(y_{r,j}^{(i)})  }{\partial y^2} u^{(i)}(y_{r,j}^{(i)})   \right|^2.
\end{align}
Here,  $\bm{u}^{(i)} = [u(x_1), u(x_2), \dots, u(x_m)]$ represents an input function evaluated at a fixed equi-spaced grid $\{x_i\}_{i=1}^m$ in $[0,1]$. For every $i$,  $\{y_{b,j}\}_{j=1}^P$ and $\{y_{r,j}\}_{j=1}^Q$ are uniformly sampled from the boundary $\partial \Omega$ and the computational domain $\Omega$ for enforcing the boundary condition and the PDE constraint, respectively. To generate a set of training data, we take $N=5 \times 10^4, P=2, Q=m = 100$ and sample $N$ input functions $\{u^{(i)}\}_{i=1}^N$ from a Gaussian random field (GRF) with a length scale $l=0.2$. We obtain a test data-set by sampling another $N=10^3$ input functions from the same GRF and solving the Poisson equation using a finite difference method on a uniform grid.

\begin{figure}
     \centering
     \begin{subfigure}[b]{0.8\textwidth}
         \centering
    \includegraphics[width=1.0\textwidth]{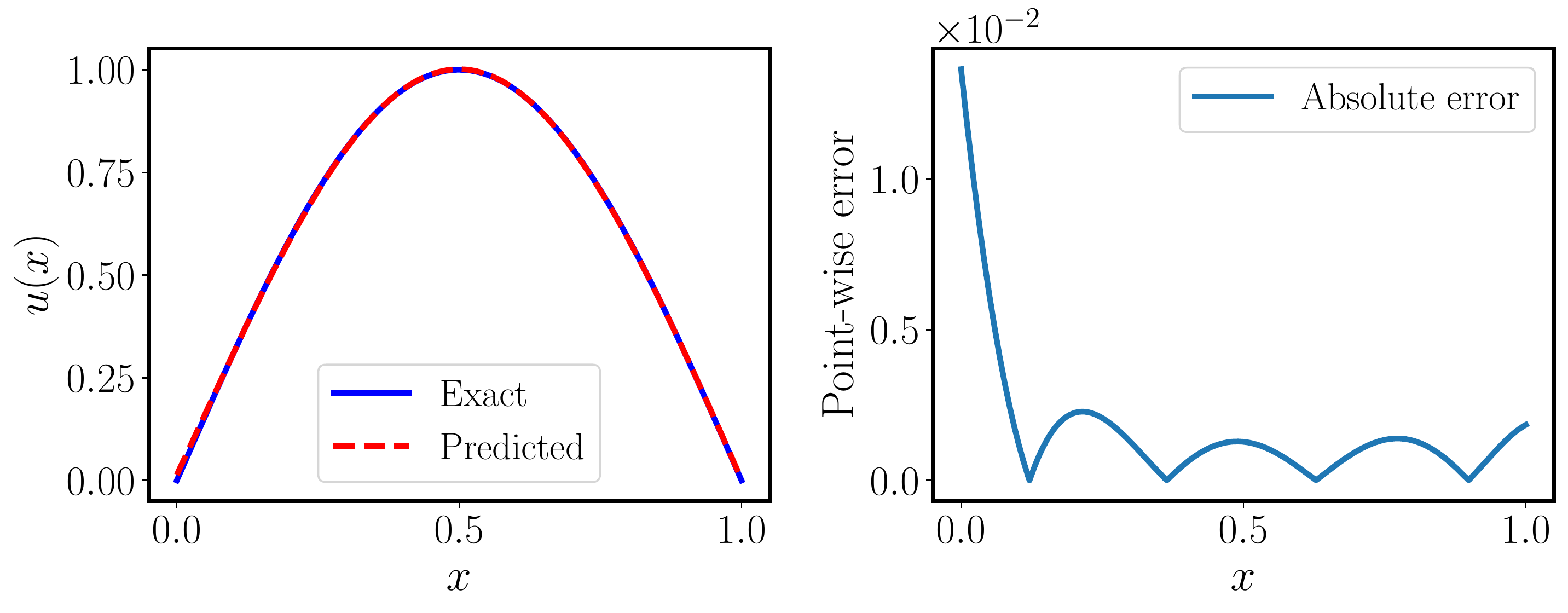}
    \label{fig: Poisson_u}
     \end{subfigure}
        \begin{subfigure}[b]{0.8\textwidth}
          \centering
    \includegraphics[width=1.0\textwidth]{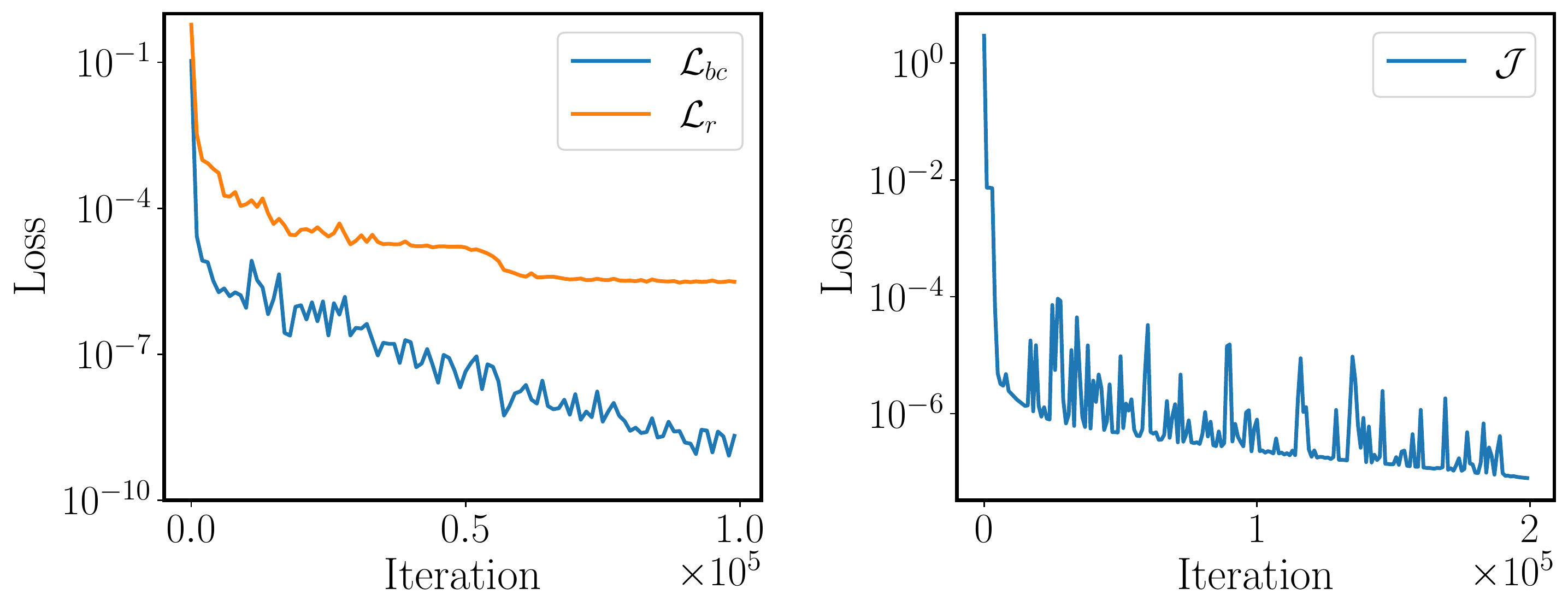}
    \label{fig: Poisson_loss}
     \end{subfigure}
      \caption{{\em Optimal control of a 1D Poisson problem:} {\em Top:} Exact control versus the predicted control. The relative $L^2$ error is  $0.37\%$.
      {\em Bottom left:} Training loss convergence of a physics-informed DeepONet for $10^5$ gradient descent iterations. {\em Bottom right:} Cost functional convergence  for $2 \times 10^5$ iterations.}
        \label{fig: Poisson}
\end{figure}

We train the physics-informed DeepONet by minimizing the above loss function for $10^5$ iterations of gradient descent using Adam optimizer \cite{kingma2014adam}. The averaged relative $L^2$ error over the test data-set is $0.09\%$. This is consistent with the visualizations shown in Figure \ref{fig: Poisson_pred}, from which an excellent agreement can be observed between the reference and the predicted solutions for different input samples in the test data-set. Then, according to step 2 in the proposed Algorithm \ref{alg: alg}, we parameterize the source term $u$ by a 5-layer fully-connected neural network $u_{\bm{\alpha}}$ with 100 neurons per hidden layer. The optimal control can be solved by minimizing the following loss function
\begin{align}
    \mathcal{J}(u_{\bm{\alpha}}) = \frac{1}{2} \int_\Omega (G_{\bm{\theta}^*}(\bm{u}_{\bm{\alpha}})(y) - d(y)) dy,
\end{align}
where $G_{\bm{\theta}^*}$ denotes the trained physics-informed DeepONet and $\bm{u}_{\bm{\alpha}} = [u_{\bm{\alpha}}(x_1), u_{\bm{\alpha}}(x_2), \dots, u_{\bm{\alpha}}(x_m) ]$ is an $m$-dimensional vector obtained by evaluating the network $u_{\bm{\alpha}}$ at the fixed sensors $\{x_i\}_{i=1}^m$. We minimize the loss function $\mathcal{J}$ for $2 \times 10^5$ iterations of gradient descent using the Adam optimizer.  The top panel of Figure \ref{fig: Poisson} presents a comparison of the learned control $u$ against the reference optimal control. It can be observed that the learned control and the ground truth are in excellent agreement with a resulting relative $L^2$ error of $0.37\%$.  Moreover, the evolution of the loss functions during training are provided in the bottom panel of Figure \ref{fig: Poisson}. It worth emphasizing that, given a trained  physics-informed DeepONet, it takes less just a few seconds to run $2 \times 10^5$ iterations of gradient descent when training the network $u_{\bm{\alpha}}$, since the PDE solution and its gradients with respect to $\bm{\alpha}$ can be evaluated rapidly. Notice, that $\bm{\alpha}$ here contains all the weights and biases parametrizing the continuous control function $u$, thus defining a $\mathcal{O}(10^4)$-dimensional PDE-constrained optimization task that is solved in a matter of seconds on a single graphics processing unit (GPU) (see Appendix \ref{app: computational cost} for a more detailed discussion on computational cost).


\subsection{Optimal control of a 2D heat equation}\label{sec:2DHeat}

Our next example aims to solve a optimal control problem involving time-dependent heat transfer. Specifically, we consider  minimizing the following objective function
\begin{align}
   \label{equ:Heat_J}
    J(u) =\int_{0}^{T} \int_{\Omega}(s-d)^{2} \mathrm{~d} \Omega \mathrm{d} t+\frac{\alpha}{2} \int_{0}^{T} \int_{\Omega} \dot{u}^{2} \mathrm{~d} \Omega \mathrm{d} t,
\end{align}
where the second integral is a regularization term that enforces smoothness of the control in time.  In particular,  the time-dependent control function $u = u(t)$ satisfies 
\begin{align}
\label{eq: heat_pde}
\frac{\partial s}{\partial t}-\nu \nabla^{2} s=u & \text { in } \Omega \times(0, T) \\
s=0 & \text { for } \Omega \times\{0\}, \\
s=0 & \text { for } \partial \Omega \times(0, T),
\end{align}
In this example we set $T = 2$, $\nu=0.01$,  $\Omega = [0, 1]^2$ and
\begin{align}\label{equ:Heat_d}
    d(x, y) = 16 xy(x-1)(y-1)\sin(\pi t).
\end{align}

Similar to the first example, we proceed by training a physics-informed DeepONet to learn the solution operator that maps source terms $u$ to the corresponding PDE solution $s$. To this end, we proceed by approximating the operator by a 7-layer modified DeepONet  $G_{\bm{\theta}}$ (see Appendix \ref{app: hp_settings},\ref{app: modified_deeponet} for details). Then, we can define the PDE residual according to equation  \ref{eq: heat_pde} as
\begin{align}
    \mathcal{R}_{\bm{\theta}}(\bm{u})(x, y, t) = \frac{\partial G_{\bm{\theta}}(x, y, t)}{\partial t} - \nu \left( \frac{\partial^2 G_{\bm{\theta}}(x, y, t)}{\partial x^2}  + \frac{\partial^2 G_{\bm{\theta}}(x, y, t)}{\partial y^2}  \right).
\end{align}
The training loss function is given as 
\begin{align}
    \label{eq: loss_heat}
    \mathcal{L}(\bm{\theta}) = \mathcal{L}_{\text{IC}}(\bm{\theta}) + \mathcal{L}_{\text{BC}}(\bm{\theta}) +  \mathcal{L}_{\text{PDE}}(\bm{\theta}), 
    \end{align}
where
\begin{align}
    &\mathcal{L}_{\text{IC}}(\bm{\theta}) = \frac{1}{NP} \sum_{i=1}^N  \sum_{j=1}^P \left| G_{\bm{\theta}}(\bm{u}^{(i)})(x_{ic, j}^{(i)}, y_{ic, j}^{(i)}, t_{ic, j}^{(i)})   \right|^2, \\
    &\mathcal{L}_{\text{BC}}(\bm{\theta}) = \frac{1}{NP} \sum_{i=1}^N  \sum_{j=1}^P \left| G_{\bm{\theta}}(\bm{u}^{(i)})(x_{bc, j}^{(i)}, y_{bc, j}^{(i)}, t_{bc, j}^{(i)})    \right|^2, \\
    &\mathcal{L}_{\text{PDE}}(\bm{\theta}) = \frac{1}{NQ} \sum_{i=1}^N  \sum_{j=1}^Q \left| \mathcal{R}_{\bm{\theta}}(\bm{u}^{(i)})(x_{r, j}^{(i)}, y_{r, j}^{(i)}, t_{r, j}^{(i)})  - u^{(i)} (t_{r, j}^{(i)}) \right|^2.
\end{align}
Here $\bm{u}^{(i)}= [u(t_1), u(t_2), \dots, u(t_m)]$ denotes an input sample evaluated at a fixed uniform grid $\{t_i\}_{i=1}^m$ in $[0,2]$. For each $i$, $\{x_{ic, j}^{(i)}, y_{ic, j}^{(i)}, t_{ic, j}^{(i)}\}_{j=1}^P$, and $\{x_{bc, j}^{(i)}, y_{bc, j}^{(i)}, t_{bc, j}^{(i)}\}_{j=1}^P$ are randomly sampled from $\Omega \times\{0\}$ and $\partial \Omega \times(0, T)$, respectively. In addition, $\{x_{r, j}^{(i)}, y_{r, j}^{(i)}, t_{r, j}^{(i)}\}_{j=1}^Q$ is a set of collocation points randomly selected from $\Omega \times(0, T)$ for enforcing the underlying PDE constraint. Specifically, we take $N = 10^3$ and $m = 100, P = Q = 100$. All input functions are sampled from a GRF with a length scale $l=0.2$.

\begin{figure}
    \centering
    \includegraphics[width=0.8\textwidth]{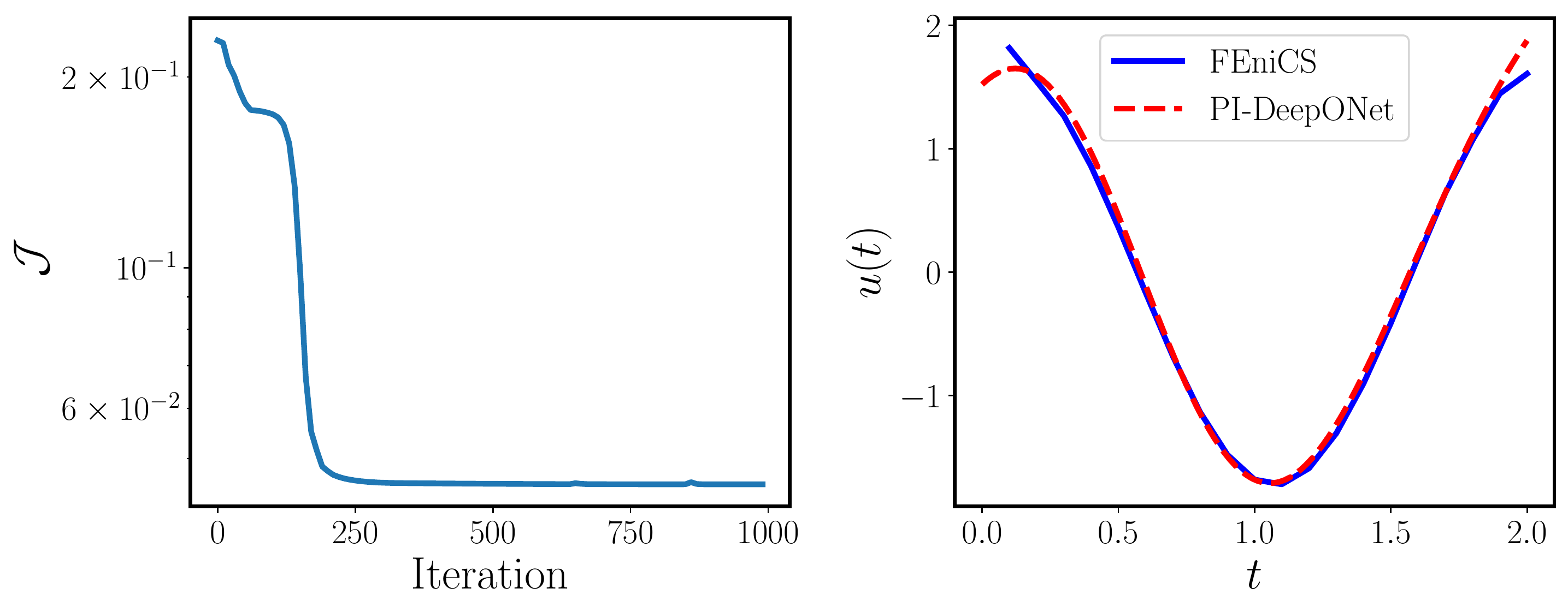}
    \caption{{\em: Optimal control of a 2D heat equation:} {\em Left:} Training cost convergence for $10^3$ iterations. {\em Right:} Inferred control function versus the baseline solution obtained  using the FEniCS finite element solver. The relative $L^2$ error is 7.44\%. An animation of the control optimization process is provided at \url{https://github.com/PredictiveIntelligenceLab/PDE-constrained-optimization-PI-DeepONet/tree/main/animations/heat}.}
    \label{fig: heat_control}
\end{figure}

\begin{figure}
    \centering
    \includegraphics[width=0.8\textwidth]{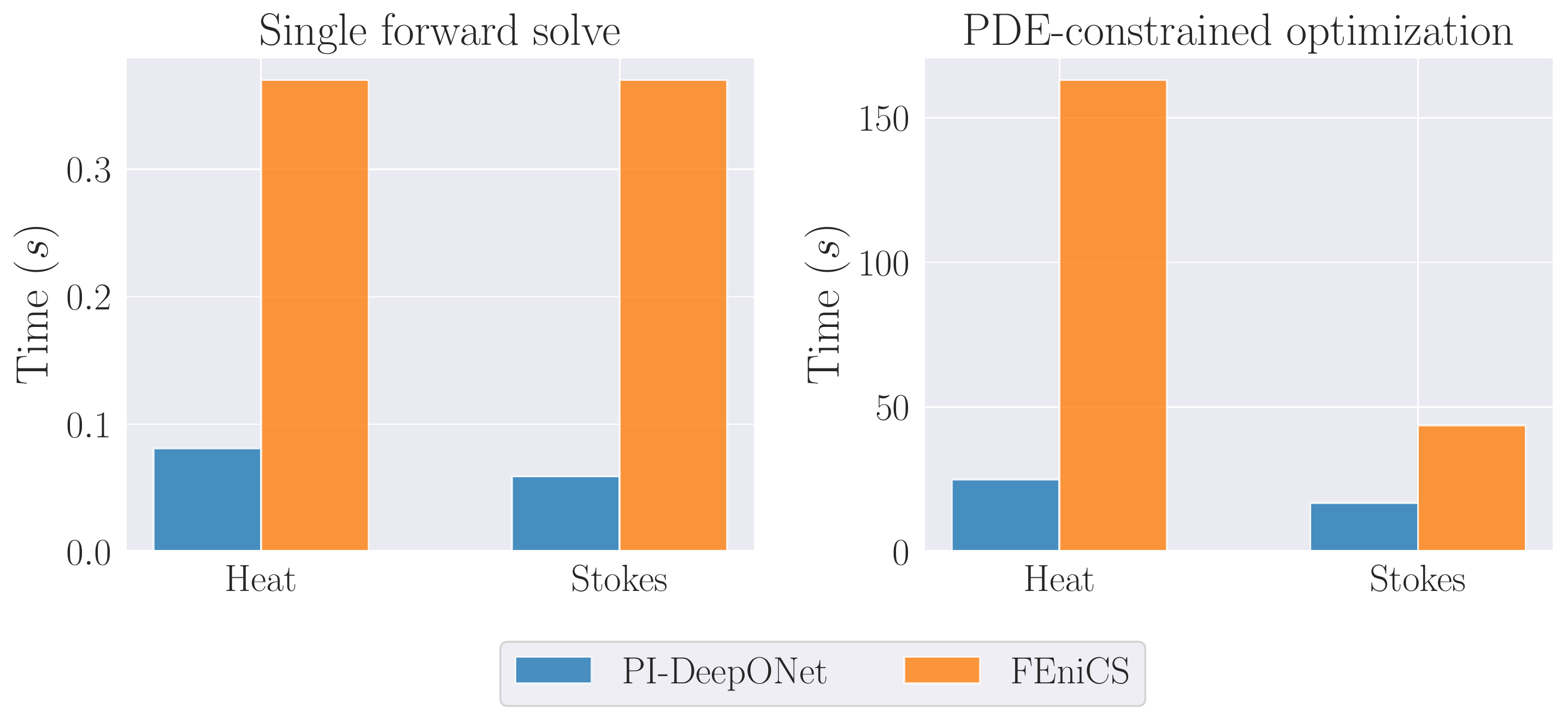}
    \caption{{\em Computational cost:} {\em Left:} Computational cost (sec) for performing inference with a trained physics-informed DeepONet model, versus solving a single PDE instance (for the 2D heat equation and Stokes flow, respectively) with the FEniCS finite element solver. {\em Right:} Training time (sec) of performing PDE-constrained optimization via the trained physics-informed DeepONet, versus a baseline adjoint method using FEniCS.}
    \label{fig: computational_costs}
\end{figure}

We train the physics-informed DeepONet by minimizing the loss of equation \ref{eq: loss_heat} for $4 \times 10^5$ iterations of gradient descent using the Adam optimizer. The resulting relative test error is $1.51\%$ averaged over all examples in the test data-set, while the loss convergence is visualized in  Figure \ref{fig: heat_deeponet_loss} (see Appendix \ref{app:validation 2d heat} for more details on validation).  Some visualizations of the predicted solution for different source terms are provided in Figure \ref{fig: heat_examples}. 
Recall that our goal is to find the desired source term that resulting in the given temperature distribution over time. We proceed by approximating the latent function $u$ by a 5-layer fully-connected neural network $u_{\bm{\alpha}}$ with 100 neurons per hidden layer. Then, this optimal control task be solved by minimizing the following cost functional
\begin{align}
    \mathcal{J}(u_{\bm{\alpha}}) = \int_0^T \int_\Omega \left(G_{\bm{\theta}^*})(\bm{u}_{\bm{\alpha}})  - d \right) d \Omega dt,
\end{align}
where $G_{\bm{\theta}^*}$ denotes the trained physics-informed DeepONet, and $\bm{u}_{\bm{\alpha}} = [u_{\bm{\alpha}}(t_1), u_{\bm{\alpha}}(t_2), \dots, u_{\bm{\alpha}}(t_m)]$ is an $m$-dimensional vector evaluated at the fixed sensors $\{t_i\}_{i=1}^m$. It is worth noting that we take $\alpha = 0$ in equation  \ref{equ:Heat_J} since the smoothness is implicitly guaranteed by the assumed neural network parametrization for $u(t)$, and therefore the regularization term is not needed. We train the network $u_{\bm{\alpha}}$ by minimizing the above cost function $\mathcal{J}$ for $1,000$ iterations. As shown in Figure \ref{fig: heat_control} and Figure \ref{fig: computational_costs}, the objective function rapidly converges after the first few hundred iterations and the total computation cost is \~25 seconds, which is 5x faster than a conventional adjoint method using the finite element solver FEniCS \cite{alnaes2015fenics} (see Appendix \ref{app: computational cost} for a more detailed discussion on computational cost). A visual assessment of the learned control is presented in the right panel of Figure \ref{fig: heat_control}, indicated that the inferred control function achieves a good agreement with the reference solution, although some discrepancies can be observed near the two end points. This may be because, in the case of the adjoint finite element solver, the control function is parametrized by a vector of $100$ discrete values, whereas in the proposed deep learning approach, the control function is a smooth continuous function parametrized by an MLP network with $\mathcal{O}(10^4)$ trainable parameters.

\begin{figure}
    \centering
    \includegraphics[width=0.4\textwidth]{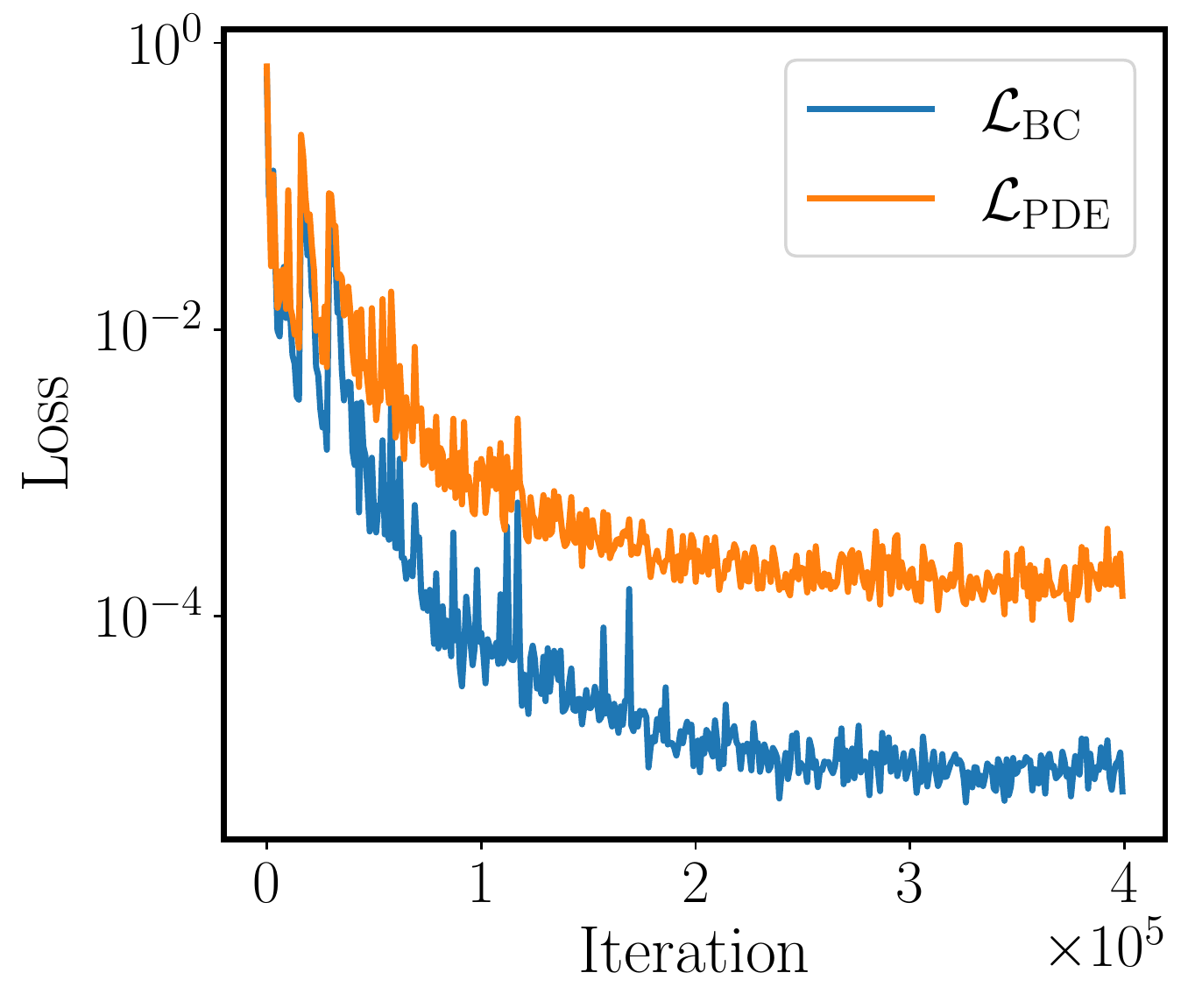}
    \caption{{\em: Optimal control of a 2D heat equation:} Training loss convergence of a physics-informed DeepONet for $4 \times 10^5$ iterations. Here we remark that we treat the initial condition as a special type of boundary condition, and thus $\mathcal{L}_{\text{IC}}$ is absorbed in $\mathcal{L}_{\text{BC}}$.}
    \label{fig: heat_deeponet_loss}
\end{figure}

\subsection{Drag minimization of obstacles in Stokes-flow}\label{sec:Stokes}

To demonstrate the capability of the proposed algorithm to perform shape optimization, our last example considers  minimizing the drag of a bluff body subject to Stokes flow. The goal is to determine the shape of an obstacle $\Gamma$, which minimizes the dissipated power in the fluid. More specifically, the cost function $\mathcal{J}$ is described as follows
\begin{align}
     \mathcal{J}(\Gamma, \bm{u}) &= \mathcal{J}_D(\Gamma) + \alpha \mathcal{J}_V(\Gamma) + \beta \mathcal{J}_B(\Gamma) \\
     & = \int_{\Omega / \Gamma} \left(\frac{\partial u}{\partial x}\right)^2 +  \left(\frac{\partial u}{\partial y}\right)^2 + \left(\frac{\partial v}{\partial x}\right)^2 + \left(\frac{\partial v}{\partial y}\right)^2 \mathrm{~d} x \mathrm{~d} y  \\
     &+ \alpha\left|\operatorname{Vol}(\Gamma)-\operatorname{Vol}\left(\Gamma_0\right)\right|^{2}  \\
     &+  \beta \sum_{j=1}^{2}\left|\mathrm{Bc}(\Gamma)-\mathrm{Bc}\left(\Gamma_{0}\right)\right|^{2},
\end{align}
where $\Omega = [0, 1] \times [0, 1]$ denotes the computational domain, $\Gamma_0$ denotes the obstacle boundary, and $\operatorname{Vol}\left(\Gamma\right)$ and $\mathrm{Bc}_{j}(\Gamma)$ is the volume and  $j$-th component of the barycenter of the obstacle, respectively. In particular,  $\bm{u} = (u, v)$ represents a velocity field subject to the Stokes equations
\begin{align}
    \label{eq: stokes_1}
    -\Delta \bm{u} + \nabla p &= 0, \quad (x,y) \in \Omega / \Gamma,    \\
    \label{eq: stokes_2}
    \nabla \cdot \bm{u}    &= 0, \quad (x,y) \in \Omega / \Gamma,
\end{align}
subject to boundary conditions
\begin{align}  
    \label{eq: stokes_3}
    \bm{u} &= \bm{0}, \quad (x,y) \in \Lambda_1 \cup \partial \Gamma, \\
        \label{eq: stokes_4}
    \bm{u} &= \bm{g}, \quad (x,y) \in \Lambda_2 \\
    \label{eq: stokes_5}
        p  & = 0,     \quad (x,y) \in \Lambda_3,
\end{align}
where $\Lambda_1$ denotes wall boundaries,  while $ \Lambda_2$ and $\Lambda_3$ correspond to the inlet and the outlet of a 2D channel, respectively.

In this example, we parametrize the shape of the obstacle $\Gamma$ by a family of ellipsoids centered at $(\frac{1}{2}, \frac{1}{2})$, i.e.
\begin{align}
    \label{eq: curve}
    \partial \Gamma  =  \partial \Gamma(\phi) = (a \cos(\phi) + \frac{1}{2}, b \sin(\phi) + \frac{1}{2}), \quad \phi \in [0, 2\pi),
\end{align}
where $a, b > 0$ denote the length of the long or the short axis. Specifically, the shape of $ \Gamma_0$ is initialized as a circle with radius $r = 0.12$. Note that an ellipse  $\Gamma$ is symmetric and its volume can be simply computed by $\operatorname{Vol}(\Gamma) = \pi a b$. Consequently $\mathcal{J}_B = 0$, and then we can exactly enforce the volume constraint $\mathcal{J}_V$ by setting $b = \frac{\operatorname{Vol}(\Gamma_0) }{\pi a }$. As such, the cost functional function for the dissipated power can be written as
\begin{align}
   \label{eq: stokes_cost}
    \mathcal{J}(\Gamma) = \mathcal{J}(a) = \int_{\Omega / \Gamma} \left(\frac{\partial u}{\partial x}\right)^2 +  \left(\frac{\partial u}{\partial y}\right)^2 + \left(\frac{\partial v}{\partial x}\right)^2 + \left(\frac{\partial v}{\partial y}\right)^2 \mathrm{~d} x \mathrm{~d} y,
\end{align}
where
\begin{align}
    \partial \Gamma  =  \partial \Gamma(\phi) = (a \cos(\phi) + \frac{1}{2}, \frac{\operatorname{Vol}(\Gamma_0) }{\pi a } \sin(\phi) + \frac{1}{2}), \quad \phi \in [0, 2\pi).
\end{align}

To represent the solution operator $G$ that maps different obstacle geometries to the associated Stokes flow solution, we employ a 7-layer modified DeepONet architecture  $G_{\bm{\theta}}$ with 100 neurons per hidden layer and with 3-dimensional vector-valued outputs for representing $u, v, p$ respectively, i.e,
\begin{align}
    \partial \bm{\Gamma}  \overset{G_{\bm{\theta}}} {\longrightarrow} [G^{(u)}_{\bm{\theta}}, G^{(v)}_{\bm{\theta}}, G^{(p)}_{\bm{\theta}}].
\end{align}
Then, we can define the following PDE residuals
\begin{align}
     \mathcal{R}_{\bm{\theta}}^{(1)}(\partial  \bm{\Gamma})(x, y) &= - \frac{\partial^2 G_{\bm{\theta}}^{(u)}( \partial \bm{\Gamma})(x, y) }{\partial x^2} - \frac{\partial^2 G_{\bm{\theta}}^{(u)}(\partial \bm{\Gamma})(x, y) }{\partial y^2} + \frac{\partial G_{\bm{\theta}}^{(p)}(\partial \bm{\Gamma})(x, y) }{\partial x}, \\
          \mathcal{R}_{\bm{\theta}}^{(2)}(\partial  \bm{\Gamma})(x, y) &= - \frac{\partial^2 G_{\bm{\theta}}^{(v)}(\partial \bm{\Gamma})(x, y) }{\partial x^2} - \frac{\partial^2 G_{\bm{\theta}}^{(v)}(\partial \bm{\Gamma})(x, y) }{\partial y^2} + \frac{\partial G_{\bm{\theta}}^{(p)}(\partial \bm{\Gamma})(x, y) }{\partial y}, \\
          \mathcal{R}_{\bm{\theta}}^{(3)}( \partial  \bm{\Gamma})(x, y) &=  \frac{\partial G_{\bm{\theta}}^{(u)}(\partial \bm{\Gamma})(x, y) }{\partial x} + \frac{\partial G_{\bm{\theta}}^{(v)}( \partial \bm{\Gamma})(x, y) }{\partial y},
\end{align}
where $\partial \bm{\Gamma}  = [\partial \Gamma(\phi_1), \partial \Gamma(\phi_2), \dots, \partial \Gamma(\phi_m)]$ denotes an input closed curve evaluated at evenly-spaced grid points $\{\phi\}_{i=1}^m$ in $[0, 2\pi]$. Then, a physics-informed DeepONet can be trained by minimizing the following weighted loss function
\begin{align}
    \label{eq: Stokes_loss}
    \mathcal{L}(\bm{\theta}) = \sum_{k=1}^3 \lambda^{(k)}_{\text{BC}} \mathcal{L}^{(k)}_{\text{BC}}(\bm{\theta}) + \sum_{k=1}^3 \lambda^{(k)}_{\text{PDE}} \mathcal{L}^{(k)}_{\text{PDE}}(\bm{\theta}),
\end{align}
where  $\{\lambda^{(k)}_{\text{BC}}\}_{k=1}^3$ and $\{\lambda^{(k)}_{\text{PDE}}\}_{k=1}^3$ are some hyper-parameters that calibrate the learning rate of each term contributing to the total loss $\mathcal{L}(\bm{\theta})$. In practice, we update these hyper-parameters using the adaptive NTK-guided weight algorithm put forth by Wang {\it et al.} \cite{wang2021improved} at each iteration during training.
Particularly,
\begin{align}
    \mathcal{L}^{(1)}_{\text{BC}}(\bm{\theta}) &= \frac{1}{NP} \sum_{i=1}^N \sum_{j=1}^{P}  \left[   \left| G^{(u)}_{\bm{\theta}}(\partial \bm{\Gamma}^{(i)}) (x_{\text{bc1}, j}^{(i)}, y_{\text{bc1},j}^{(i)}) \right|^2 +   \left| G^{(v)}_{\bm{\theta}}(\partial \bm{\Gamma}^{(i)}) (x_{\text{bc1}, j}^{(i)}, y_{\text{bc1},j}^{(i)}) \right|^2 \right], \\
    \mathcal{L}^{(2)}_{\text{BC}}(\bm{\theta}) &= \frac{1}{NP} \sum_{i=1}^N \sum_{j=1}^{P}  \left[ \left| G^{(u)}_{\bm{\theta}}(\partial \bm{\Gamma}^{(i)}) (x_{\text{bc2}, j}^{(i)}, y_{\text{bc2},j}^{(i)}) - \sin(\pi y_{\text{bc2},j}^{(i)}) \right|^2 +  \left| G^{(v)}_{\bm{\theta}}(\partial \bm{\Gamma}^{(i)}) (x_{\text{bc2}, j}^{(i)}, y_{\text{bc2},j}^{(i)}) \right|^2 \right], \\
    \mathcal{L}^{(3)}_{\text{BC}}(\bm{\theta}) &= \frac{1}{NP} \sum_{i=1}^N \sum_{j=1}^{P}\left| G^{(p)}_{\bm{\theta}}(\partial \bm{\Gamma}^{(i)}) (x_{\text{bc3}, j}^{(i)}, y_{\text{bc3},j}^{(i)}) \right|^2,
\end{align}
and 
\begin{align}
   \mathcal{L}^{(k)}_{\text{PDE}}(\bm{\theta}) = \frac{1}{NQ} \sum_{i=1}^N \sum_{j=1}^{Q}  \left|  \mathcal{R}_{\bm{\theta}}^{(k)}(\partial \bm{\Gamma}^{(i)})(x^{(i)}_{\text{r}, j}, y^{(i)}_{\text{r}, j})   \right|^2, \quad k = 1,2,3.
\end{align}
Here, for each input sample $\partial \bm{\Gamma}^{(i)}$, $\{x_{\text{bc1}, j}^{(i)}, y_{\text{bc1},j}^{(i)}  \}_{j=1}^P$, $\{x_{\text{bc2}, j}^{(i)}, y_{\text{bc2},j}^{(i)}  \}_{j=1}^P$, $\{x_{\text{bc3}, j}^{(i)}, y_{\text{bc3},j}^{(i)}  \}_{j=1}^P$ and $\{x^{(i)}_{\text{r}, j}, y^{(i)}_{\text{r}, j} \}_{j=1}^Q$ are uniformly sampled at the boundaries $\Lambda_1 \cup \partial \Gamma$, $\Lambda_2$, $\Lambda_3$  and inside the domain $\Omega / \Gamma$, respectively.  In this example, we set $N = 1000, m=P=100, Q=2500$. To prepare the training data-set, we randomly sample  $a, b$ from a uniform distribution $\mathcal{U}(0.05, 0.3)$, and obtain a set of input curves $\{\partial \bm{\Gamma}^{(i)} \}_{i=1}^N$ using equation (\ref{eq: curve}). To generate a set of test data, we repeat the same procedure to obtain $500$ new curves, and obtain the corresponding PDE solutions using the FEniCS solver \cite{alnaes2015fenics} (see validation details in Appendix \ref{app: validation stokes}).  

\begin{figure}
    \centering
    \includegraphics[width=0.4\textwidth]{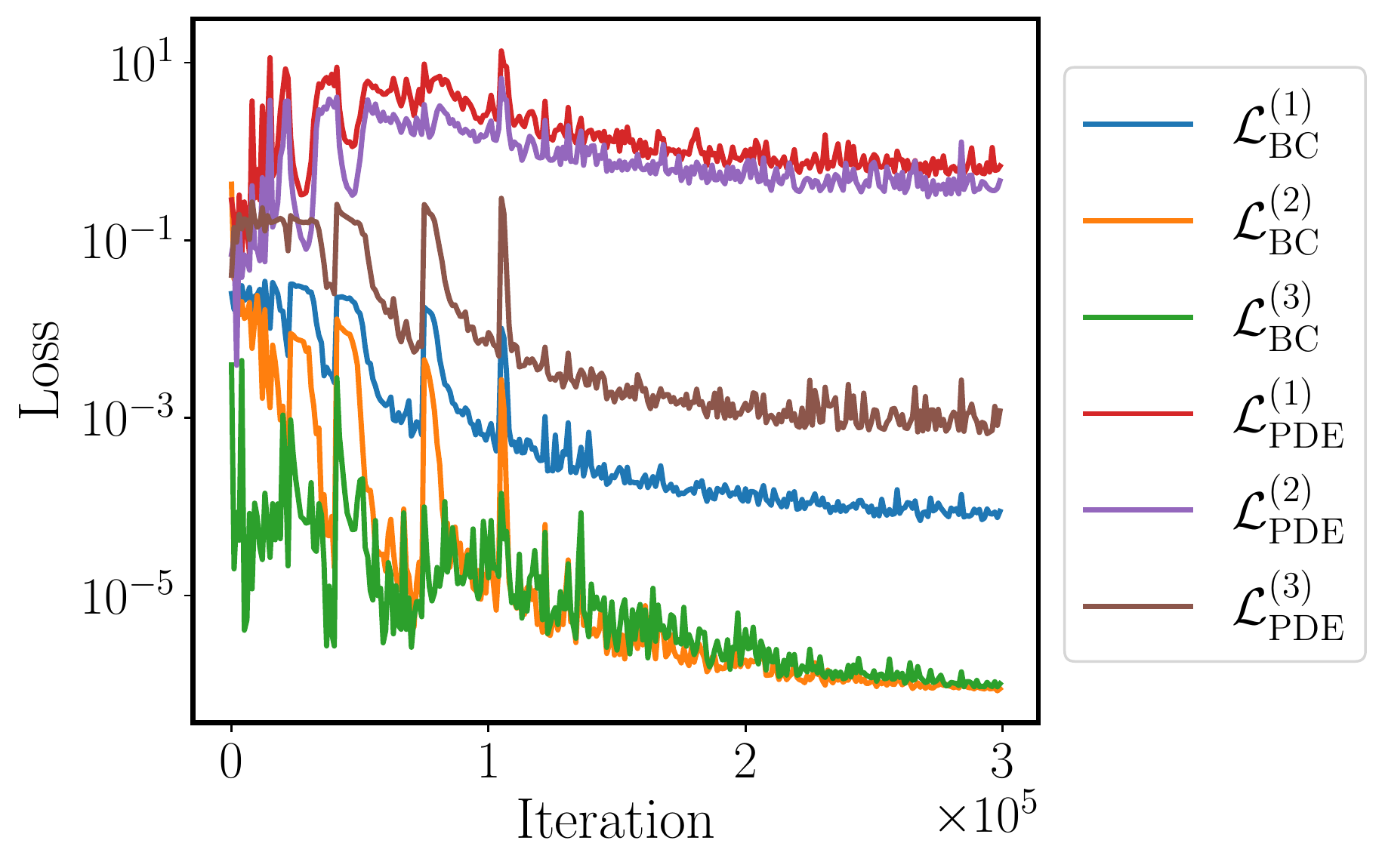}
 \caption{{\em: Drag minimization of obstacles in Stokes-flow:} Training loss convergence of a physics-informed DeepONet for $3 \times 10^5$ iterations.}
    \label{fig: stokes_deeponet_loss}
\end{figure}

We train the physics-informed DeepONet for $3 \times 10^5$ iterations of gradient descent using the Adam optimizer. The loss curves and the resulting test error are shown in Figure \ref{fig: stokes_deeponet_loss} and Table \ref{tab: stokes_error}, respectively. Moreover, some visualizations of the predicted solution across different obstacle shapes are provided in Figure \ref{fig: Stokes_PI_deeponet_modified_deeponet_global_NTK_weights_pred_1} - \ref{fig: Stokes_PI_deeponet_modified_deeponet_global_NTK_weights_pred_14}.  From these results, one may conclude that the trained physics-informed DeepONet  approximates the solution operator with sufficient accuracy. 

Next, we can employ the trained physics-informed DeepONet to expediently minimize the cost functional of equation \ref{eq: stokes_cost} as
\begin{align}
   \mathcal{J}(a) = \int_{\Omega / \Gamma} & \left(\frac{\partial G_{\bm{\theta^*}}^{(u)}(\partial \bm{\Gamma}) (x, y )  }{\partial x}\right)^2 +  \left(\frac{\partial  G_{\bm{\theta^*}}^{(u)}(\partial \bm{\Gamma}) (x, y ) }{\partial y}\right)^2 \\
   +&\left(\frac{\partial  G_{\bm{\theta^*}}^{(v)}(\partial \bm{\Gamma}) (x, y ) }{\partial x}\right)^2 + \left(\frac{\partial  G_{\bm{\theta^*}}^{(v)} (\partial \bm{\Gamma}) (x, y ) }{\partial y}\right)^2 \mathrm{~d} x \mathrm{~d} y,
\end{align}
where $G_{\bm{\theta^*}}$ denotes the trained physics-informed DeepONet. Notice here that parametrized the obstacle shape $\partial \bm{\Gamma}  = [\partial \Gamma(\phi_1), \partial \Gamma(\phi_2), \dots, \partial \Gamma(\phi_m)]$ using $m=100$ evenly-spaced grid points $\{\phi\}_{i=1}^m$ in $[0, 2\pi]$. It is also worth noting that all  gradients required to evaluate the above cost functional and its derivatives can be computed via automatic differentiation.

\begin{figure}
    \centering
    \includegraphics[width=0.8\textwidth]{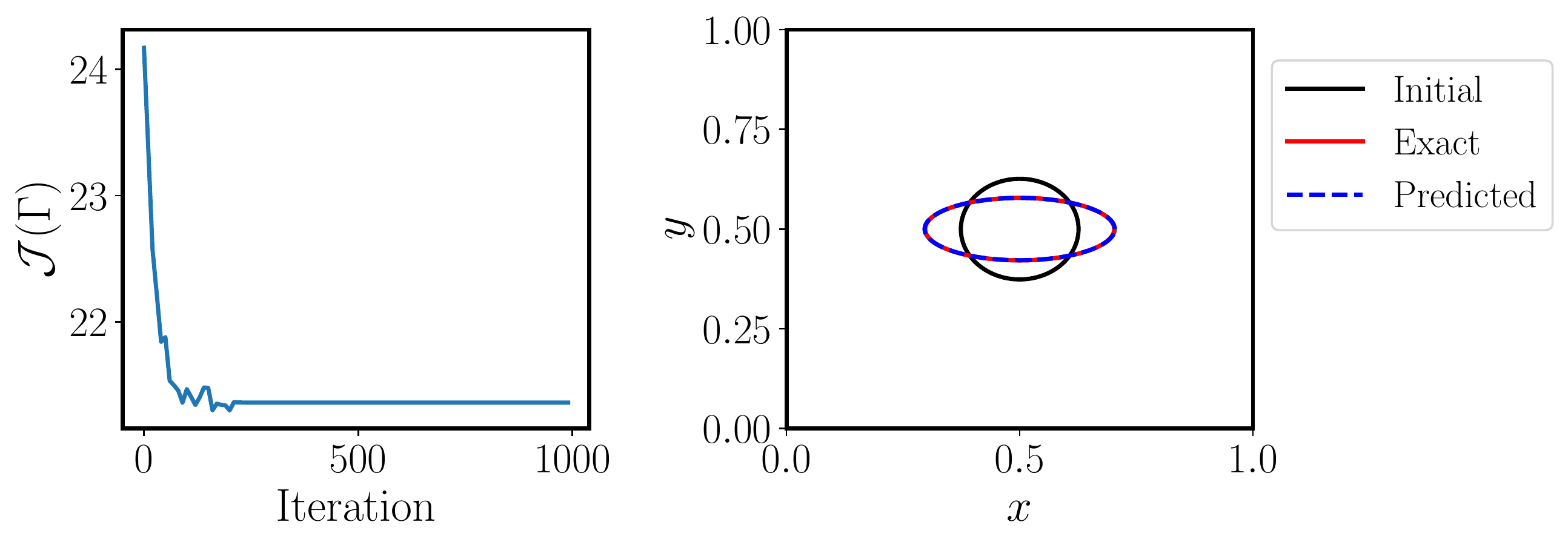}
    \caption{{\em Drag minimization of obstacles in Stokes-flow:} {\em Left:} Convergence of the target cost functional for $10^3$ gradient descent iterations. {\em Right:} Learned optimal obstacle geometry versus the reference solution obtained by the adjoint finite element solver FEniCS. The relative $L^2$ error is 0.51\%.  An animation of the shape optimization process can be found in \url{https://github.com/PredictiveIntelligenceLab/PDE-constrained-optimization-PI-DeepONet/tree/main/animations/Stokes}.}
    \label{fig: Stokes_J_obstacle}
\end{figure}

The left panel of Figure  \ref{fig: Stokes_J_obstacle} shows the convergence of the cost functional during minimization via gradient descent. We can see that $\mathcal{J}(a)$ decreases rapidly and converges to a local minimum within a few hundred iterations. Moreover, the overall optimization time takes less than 20s, which is noticeably faster than employing an adjoint finite element solver as a forward model in this loop, see figure \ref{fig: computational_costs}.
From the right panel of Figure  \ref{fig: Stokes_J_obstacle}, the predicted  optimal obstacle achieves an excellent agreement with the reference solution, with a resulting relative error of $0.51\%$ (see Appenix \ref{app: validation stokes} for more details on validation). 
Finally, we remark that a DeepONet with a conventional architecture \cite{lu2021learning} may not be able to solve this control problem,  since the cost functional requires the gradients of the predicted velocity field, which are possibly inaccurate as discussed in Wang {\it et al.} \cite{wang2021learning}.

\begin{table}
    \centering
    \renewcommand{\arraystretch}{1.4}
    \begin{tabular}{|c|c|c|c|} 
    \hline
           Latent variable  & $u$  & $v$  & $p$  \\
             \hline
       Test error   & $0.97\% \pm 0.24\%$  &$4.79\% \pm 0.53\%$  &   $1.53\% \pm 0.67\%$   \\
         \hline
    \end{tabular}
    \caption{{\em: Drag minimization of obstacles in Stokes-flow:} Test error of the predicted velocity and pressure field obtained from a trained physics-informed DeepONet.}
    \label{tab: stokes_error}
\end{table}

\section{Discussion}
\label{sec: summary}

We presented a self-supervised deep learning framework that enables the construction of fast and differentiable surrogates for solving parametric PDEs, and illustrated its effectiveness in tackling high-dimensional PDE-constrained optimization problems. The proposed physics-informed DeepONet surrogates can be trained in the absence of paired input-output data, and therefore do not require one to generate large training data-sets by repeatedly evaluating expensive simulators or experiments, as is typically the case for building reduced order models. We examined and validated the effectiveness of the proposed framework through a series of numerical experiments involving time-distributed optimal control of heat transfer, and drag minimization of Stokes flow over obstacles.  Our findings suggest that the proposed is not only capable of handling continuous, infinite-dimensional control functions, but also is notably faster than conventional adjoint solvers with comparable predictive accuracy. Given the prominence of PDE-constrained optimization problems across all corners of science and engineering, we expect that this work can have a broad technical impact in accelerating the design and control of complex systems.

Currently the main roadblock for scaling the proposed techniques to more realistic applications is the ability of physics-informed neural networks and physics-informed DeepONets to accurately solve more complex PDEs involving multi-scale solutions, stiff dynamics, and large spatio-temporal domains. Issues including spectral bias \cite{wang2021eigenvector}, ill-conditioned loss landscapes \cite{krishnapriyan2021characterizing}, and gradient pathologies \cite{wang2021understanding}, pose practical difficulties in training physics-informed deep learning models in these more challenging settings. Although progress is being made \cite{wang2020and,wang2021improved}, we are still in the early stages of demystifying how physics-informed deep learning models can be a reliable and effective tool for emulating complex PDEs in the absence of labelled training data. To this end, designing more effective neural network architectures, initialization schemes and optimizers will be the key to unlocking the full potential of physics-informed DeepONets in enabling the real-time solution of more realistic PDE-constrained optimization problems, and beyond.

\section*{Acknowledgments}
This work received support from DOE grant DE-SC0019116, AFOSR grant FA9550-20-1-0060, and DOE-ARPA grant DE-AR0001201. 
We would also like to thank the developers of the software that enabled our research, including JAX \cite{jax2018github}, Matplotlib \cite{hunter2007matplotlib}, NumPy \cite{harris2020array}, FEniCS \cite{alnaes2015fenics}, and dolfin-adjoint \cite{mitusch2019dolfin}.

\section*{Competing Interests}
The authors declare that they have no competing interests.

\section*{Author Contributions}
SW and PP conceptualized the research and designed the numerical studies. SW implemented the methods and conducted the numerical experiments. MAB assisted with the numerical studies and generated the validation data-sets. PP provided funding and supervised all aspects of this work. SW, MAB and PP wrote the manuscript.

\section*{Data and materials availability}
All methods needed to evaluate the conclusions in the paper are present in the paper and/or the Appendix. All code and data accompanying this manuscript will be made publicly available at \url{https://github.com/PredictiveIntelligenceLab/PDE-constrained-optimization-PI-DeepONet}.

\bibliographystyle{unsrt}
\bibliography{references}

\clearpage

\appendix
\clearpage
\section{Nomenclature}
Table \ref{tab: Notations} summarizes the main symbols and notation used in this work.

\begin{table}[ht]
\renewcommand{\arraystretch}{1.4}
    \centering
    \begin{tabular}{ll} 
    \Xhline{3\arrayrulewidth} 
    Notation     & Description \\
    \Xhline{3\arrayrulewidth} 
        $\mathcal{J}$ & cost function of the PDE-constrained optimization \\
        $\mathcal{N}$  & differential operator  \\
        $\mathcal{B}$  & boundary condition of a PDE system \\
         $\bm{u}(\cdot)$ & an input function \\  
        $\bm{s}(\cdot)$ & a solution to a parametric PDE \\
        $G$         & an operator \\
        $G_{\bm{\theta}}$  &  a  DeepONet representation of the operator $G$ \\
        $u_{\bm{\alpha}}$  &  a fully-connected neural network representation of the control \\
        $\bm{\theta}$ &  all trainable parameters of a DeepONet \\
         $\bm{\alpha}$ &  all trainable parameters of a control representation \\
        $\{\bm{x}_i\}_{i=1}^m$  & $m$ sensor points where input functions $\bm{u}(\bm{x})$ are evaluated\\
        $[u(\bm{x}_1), u(\bm{x}_2), \dots, u(\bm{x}_m)]$ & an input of the branch net, representing the input function $u$ \\ 
        $\bm{y}$          & an input of the trunk net, a point in the domain of $G(u)$ \\
        N                & number of input samples in the training data-set \\
        m               &  number of locations for evaluating the input functions $u$        \\
        P               &  number of locations for evaluating the output functions $G(u)$        \\
        Q                & number of collocation points for evaluating the PDE residual \\
        MLP & Multi-Layer Perceptron \\
        GRF             &  a Gaussian random field                   \\
        $l$              & length scale of a Gaussian random field   \\
    \Xhline{3\arrayrulewidth}
    \end{tabular}
    \caption{{\em Nomenclature}: Summary of the main symbols and notation used in this work.}
    \label{tab: Notations}
\end{table}

\clearpage
\section{Hyper-parameter settings}
\label{app: hp_settings}

\begin{table}[ht]
\renewcommand{\arraystretch}{1.4}
    \centering
    \begin{tabular}{c|ccccccc}
        \Xhline{3\arrayrulewidth}
      Case   & Input function space &  m & P & Q & \# u Train & \# u Test & Iterations  \\
      \hline
     1D Poisson equation & GRF$(l=0.2)$ & $10^2$ & 2 & $10^2$ & $10^4$ & $10^3$  & $ 10^{5}$ \\
    2D Heat equation & GRF$(l=0.5)$ & $10^2$ & $10^2$ & $2 \times 10^3$ & $10^3$ & $10^2$ & $4 \times 10^{5}$\\
        Stokes flow &  Ellipse & $10^2$ & $10^2$ & $2 \times 10^3$ & $  10^3$ & $10^2$ & $3 \times 10^{5}$ \\
    \Xhline{3\arrayrulewidth}
    \end{tabular}
    \caption{Default hyper-parameter settings for training physics-informed DeepONets in each benchmark (unless otherwise stated).}
    \label{tab: parameters_case}
\end{table}

\begin{table}[ht]
\renewcommand{\arraystretch}{1.4}
    \centering
    \begin{tabular}{c|ccccccc}
        \Xhline{3\arrayrulewidth}
        Case   & Architecture &  Trunk depth &  Trunk width & Branch depth & Branch width  \\
        \hline
        1D Poisson equation &  Modified DeepONet & 5 & $10^2$ & 5 & $10^2$ \\
         2D Heat equation &  Modified DeepONet & 7 & $10^2$ & 7 & $10^2$ \\
           Stokes flow &  Modified DeepONet & 7 & $10^2$ & 7 & $10^2$ \\
       \Xhline{3\arrayrulewidth}
    \end{tabular}
    \caption{Physics-informed DeepONet architectures for each benchmark employed in this work (unless otherwise stated). The details of the Modified DeepONet architecture are provided in \ref{app: modified_deeponet}.}
    \label{tab: Physics_informed_DeepONet_size}
\end{table}

\begin{table}[ht]
\renewcommand{\arraystretch}{1.4}
    \centering
    \begin{tabular}{c|ccccccc}
        \Xhline{3\arrayrulewidth}
        Case   & Architecture &   Depth &  Width & Iterations   \\
        \hline
            1D Poisson equation &  MLP & 5 & $10^2$ & $2 \times 10^5$ \\
         2D Heat equation &  MLP & 5 & $10^2$ & $10^3$  \\
           Stokes flow &  - & - & - & $10^3$ \\
       \Xhline{3\arrayrulewidth}
    \end{tabular}
    \caption{Neural network architectures for representing the control variables in each benchmark (unless otherwise stated). }
    \label{tab: network_size}
\end{table}

\clearpage
\section{Computational cost}\label{app: computational cost}

\begin{table}[ht]
\renewcommand{\arraystretch}{1.4}
    \centering
    \begin{tabular}{c|c}
     \Xhline{3\arrayrulewidth}
       Case  &   Training time (hours)  \\
     \hline
      1D Poisson equation   &  0.32 \\
      2D Heat equation  &   22.38 \\
      Stokes flow &  30.12 \\
    \Xhline{3\arrayrulewidth}
    \end{tabular}
    \caption{Computational cost (hours) for training  physics-informed DeepONet models across the different benchmarks. Reported timings are obtained on a single NVIDIA RTX A6000 graphics card.}
    \label{tab: computational_cost}
\end{table}

\section{Training procedure and error metric}
\label{app: metrics}

Throughout all benchmarks we  employ hyperbolic tangent activation functions (Tanh) and initialize the DeepOnet networks using the Glorot normal scheme \cite{glorot2010understanding}, unless otherwise stated. All networks are trained via  mini-batch stochastic gradient descent using the Adam optimizer \cite{kingma2014adam} with default settings.  Particularly, we set the batch size to be $10,000$ and use exponential learning rate decay with a decay-rate of 0.9 every $2,000$ training iterations. Moreover, we employ a modified DeepONet architecture for representing the solution operator \cite{wang2021improved}, which is empirically proved to perform better than the conventional DeepONet architecture. The forward pass of this architecture is illustrated in Appendix \ref{app: modified_deeponet}. For the Poisson and Stokes benchmarks, an NTK-guided weighting scheme is used for training physics-informed DeepONets, see \cite{wang2021improved} for more details.

The error metric employed throughout all numerical experiments to assess model performance is the relative $L^2$ norm. Specifically, the reported test errors correspond to the mean of the relative $L^2$ error of a trained model over all examples in the test data-set, i.e
\begin{align}\label{eq: test_error}
    \text{Test error} := \frac{1}{N} \sum_{i=1}^N \frac{\|G_{\theta}(\bm{u}^{(i)})(y) - G(\bm{u}^{(i)})(y)\|_2 }{\|G(\bm{u}^{(i)})(y)\|_2},
\end{align}
where $N$ denotes the number of examples in the test data-set and $y$ is typically a set of equi-spaced points in the domain of $G(u)$. Here $G_{\theta}(\bm{u}^{(i)})(\cdot)$ denotes a predicted DeepONet output function, while $G(\bm{u}^{(i)})(\cdot)$ corresponds to the ground truth target functions.

\section{Modified DeepONet architecture}
\label{app: modified_deeponet}

The forward pass of a L-layer modified DeepONet architecture \cite{wang2021improved} is given as follows
\begin{align}
     & \bm{U} = \phi( \bm{W}_u \bm{u} + \bm{b}_u), \ \  \bm{V} = \phi( \bm{W}_y \bm{y} + \bm{b}_y) \\
     & \bm{H}_u^{(1)} = \phi(\bm{W}^{(1)}_u \bm{u}  + \bm{b}^{(1)}_u), \ \ \bm{H}_y^{(1)} = \phi( \bm{W}^{(1)}_y \bm{y} + \bm{b}^{(1)}_y) \\
    & \bm{Z}_u^{(l)} = \phi(\bm{W}^{(l)}_u \bm{H}^{(l)}_u  + \bm{b}^{(l)}_u), \ \ \bm{Z}_y^{(l)} = \phi( \bm{W}^{(l)}_y \bm{H}^{(l)}_y + \bm{b}^{(l)}_y), \quad l = 1, 2, \dots, L-1 \\
    \label{eq: arch_embed_1}
    & \bm{H}^{(l+1)}_u = (1 - \bm{Z}^{(l)}_u) \odot \bm{U}  +  \bm{Z}^{(l)}_u  \odot \bm{V}, \quad l = 1, \dots, L-1 \\
     \label{eq: arch_embed_2}
    & \bm{H}^{(l+1)}_y = (1 - \bm{Z}^{(l)}_y) \odot \bm{U}  +  \bm{Z}^{(l)}_y  \odot \bm{V}, \quad l = 1, \dots, L-1 \\
    & \bm{H}_u^{(L)} = \phi(\bm{W}^{(L)}_u \bm{H}^{(L-1)}_u  + \bm{b}^{(L)}_u), \ \ \bm{H}_y^{(L)} = \phi( \bm{W}^{(L)}_y \bm{H}^{(L-1)}_y   + \bm{b}^{(L)}_y) \\
    &G_{\bm{\theta}}(\bm{u})(\bm{y}) = \left\langle \bm{H}_u^{(L)}, \bm{H}_y^{(L)}        \right\rangle,
\end{align}
where $\odot$ denotes point-wise multiplication, $\phi$ denotes a activation function, and  $\bm{\theta}$ represents all  trainable parameters of the DeepONet model. In particular, $\{\bm{W}_u^{(l)},  \bm{b}_u^{(l+1)} \}_{l=1}^{L+1}$ and $ \{\bm{W}_y^{(l)},  \bm{b}_y^{(l+1)} \}_{l=1}^{L+1} $ are the weights and biases of the branch and trunk networks, respectively. Wang {\it et al.} \cite{wang2021improved} have demonstrated that this Modified DeepONet architecture is more resilient against vanishing gradient pathologies and can consistently outperform the conventional architecture put forth by Lu {\it et al.} \cite{lu2021learning}.

\clearpage
\section{Validation}\label{app: validation}

We consider Finite Element (FE) solutions in order to validate the DeepONet results to the problem of optimal control of 2D heat equation (section \ref{sec:2DHeat}) and to the problem of drag minimization over an obstacle in Stokes-flow (section \ref{sec:Stokes}). The FE solutions are generated using the FEniCS solver \cite{alnaes2015fenics} on a 4-core MacBook Pro laptop (with a 2 GHz Intel CPU and 16 GB RAM).

\subsection{Optimal control of 2D heat equation}\label{app:validation 2d heat}

Since the 2D heat equation is a spatio-temporal PDE, we consider the forward Euler scheme for the finite difference discretization in time, and a finite element (FE) approximation for the space discretization \cite{Funke2013Fenics,Mitusch2019fenics}.

For the FE approximation, we rely on a continuous Galerkin method. In order to establish the required resolutions for the different discretization parameters, we perform a grid search over: the time step, the FE discretization order, the FE mesh size, and the tolerance of the L-BFGS algorithm. The high-fidelity solution is taken such that the time step is equal to $dt=0.01$, the mesh size is equal to $h=0.015625$, the FE discretization is of second order (i.e. $\mathbb{P}_2$ FE approximation space), and the L-BFGS tolerance is equal to $10^{-12}$. The optimal source term $u_{HF}$ obtained with these settings as a solution to the problem (\ref{equ:Heat_J})-(\ref{equ:Heat_d}) is considered as the high-fidelity solution, and we perform a grid search over the discretization parameters listed above in order to pick the resolutions that provide an optimal source term $u^*$ with a relative error below $1\%$ compared to $u_{HF}$. Such criteria provides the following discretization parameters: the time step is equal to $dt=0.02$, the mesh size is equal to $h=0.0625$, the FE discretization is of first order (i.e. $\mathbb{P}_1$ FE approximation space), and the L-BFGS tolerance is equal to $10^{-9}$.

Performing the PDE-constrained optimization with the discretization settings obtained with the grid search and specified above (which gives the optimal source term $u^*$) requires $163 \ s$ on a 4-core MacBook Pro laptop (with a 2 GHz Intel CPU and 16 GB RAM). The forward pass required to estimate the FE solution with these settings takes $0.37 \ s$. As a reference, if one opts for a finer mesh size equal to $h=0.015625$, as used to obtain the high fidelity solution $u_{HF}$, then the PDE-constrained optimization takes $605 \ s$, while the forward pass takes $2.55 \ s$.

\subsection{Drag minimization over an obstacle in Stokes-flow}\label{app: validation stokes}

For the FE approximation of the problem of drag minimization over an obstacle in Stokes-flow, the stresses $h$ at the obstacle-fluid boundary $\partial\Gamma$ are taken as the design parameters. In such a setting, the fluid domain changes from it unperturbed state $\Omega_0$ to the new one $\Omega(s)=\{x+s(h)|x\in\Omega_0\}$ where $s$ is the solution to a linear elasticity problem \cite{Schulz2016Shape,Dokken2020Fenics}. During the optimization process, the Stokes equation (\ref{eq: stokes_5})-(\ref{eq: stokes_1}) is solved by considering the changed fluid domain $\Omega(s)$ and using the stable Taylor-Hood finite element space as test and trial space. We verify that the code recovers the well characterized optimal geometry for the drag minimization over an obstacle in Stokes-flow and which consists of a rugby shaped ball with a $90$ degree front and back corners \cite{Pironneau1974}.  

Performing the PDE-constrained optimization with the FE approximation specified above requires $43.69 \ s$ on a 4-core MacBook Pro laptop (with a 2 GHz Intel CPU and 16 GB RAM). The forward pass required to estimate the FE solution with these settings takes $0.37 \ s$.

\clearpage
\section{Supplementary Visualizations}

\subsection{1D Poisson equation}

\begin{figure}[ht]
     \centering
     \begin{subfigure}[b]{0.45\textwidth}
         \centering
    \includegraphics[width=1.0\textwidth]{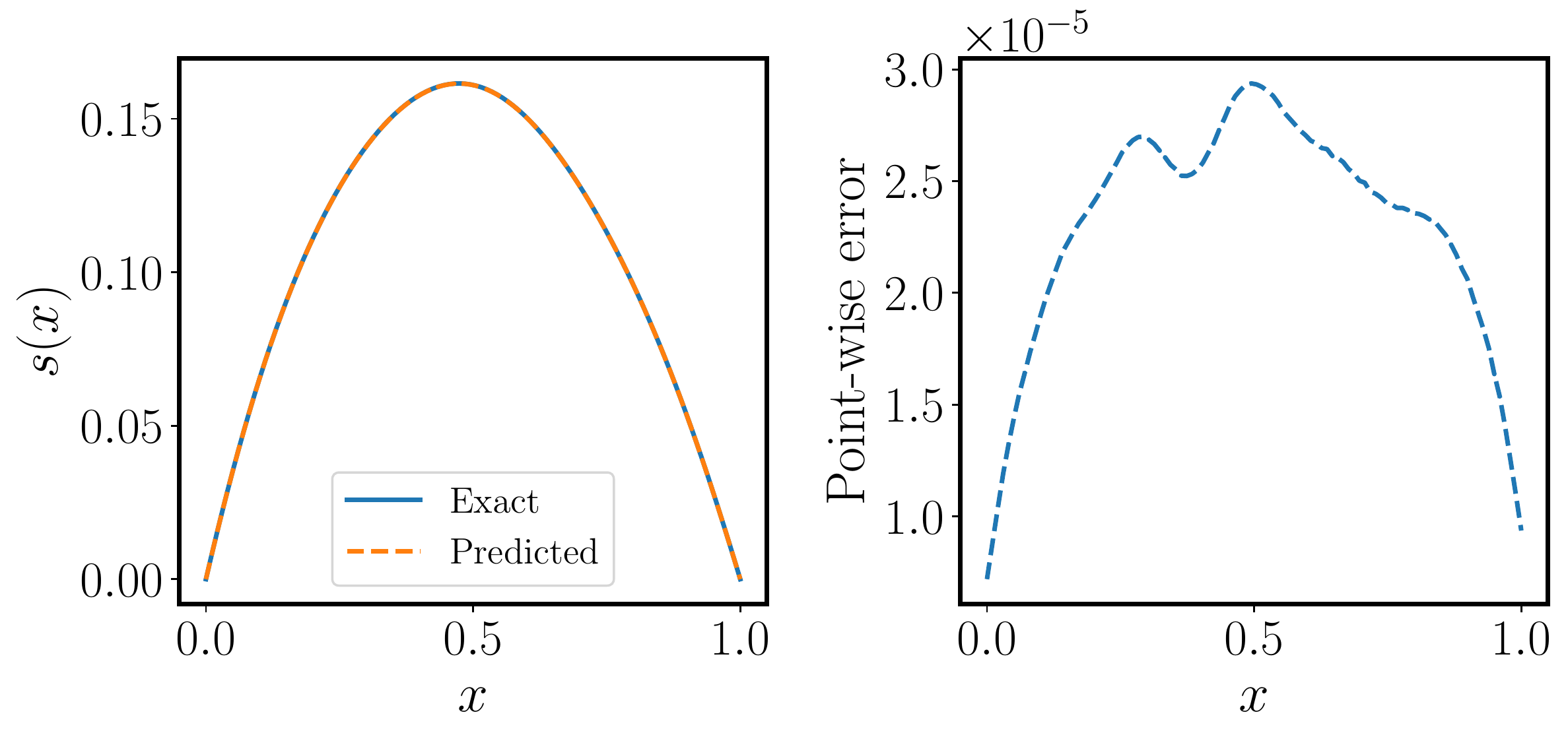}
     \end{subfigure}
            \begin{subfigure}[b]{0.45\textwidth}
         \centering
    \includegraphics[width=1.0\textwidth]{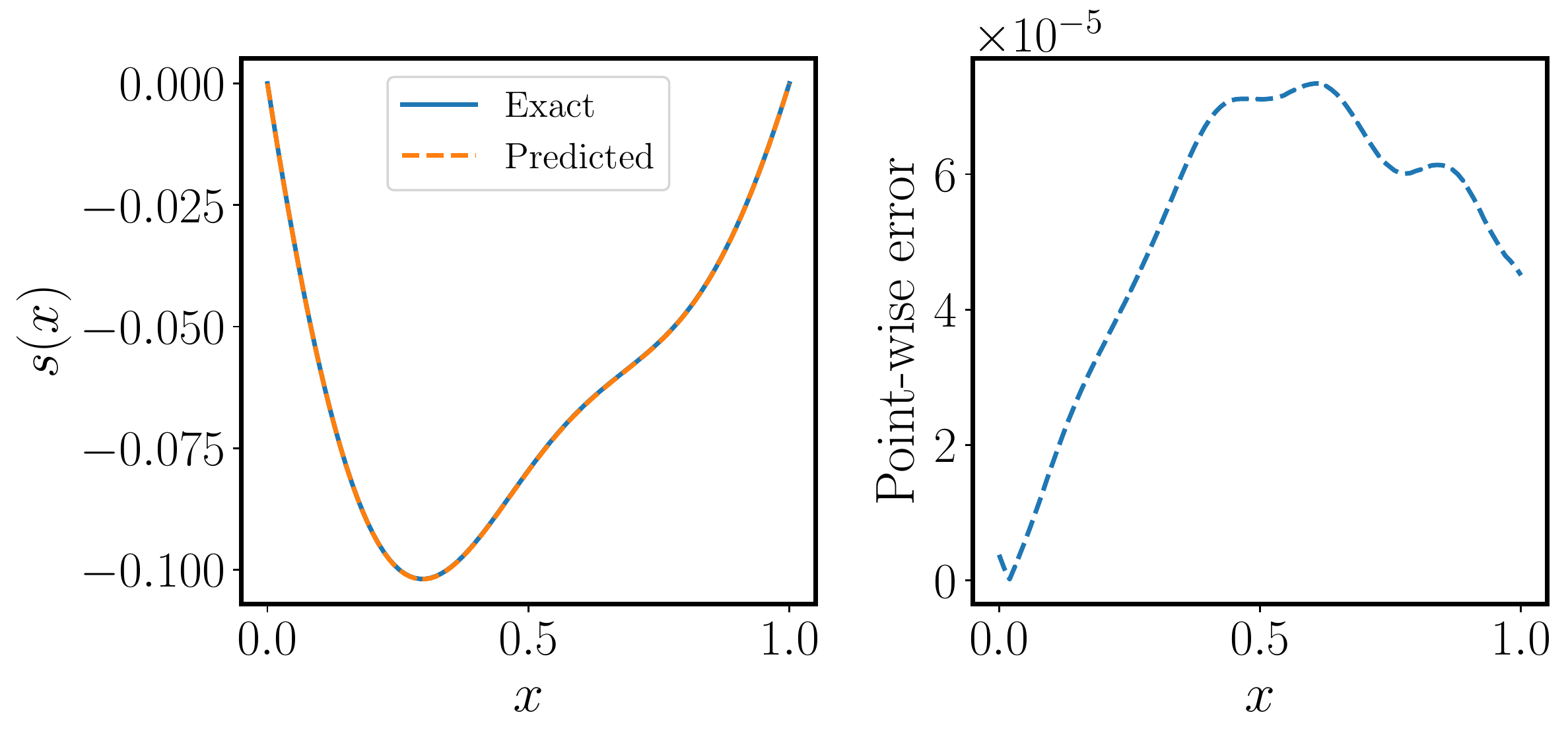}
     \end{subfigure}
          \begin{subfigure}[b]{0.45\textwidth}
         \centering
    \includegraphics[width=1.0\textwidth]{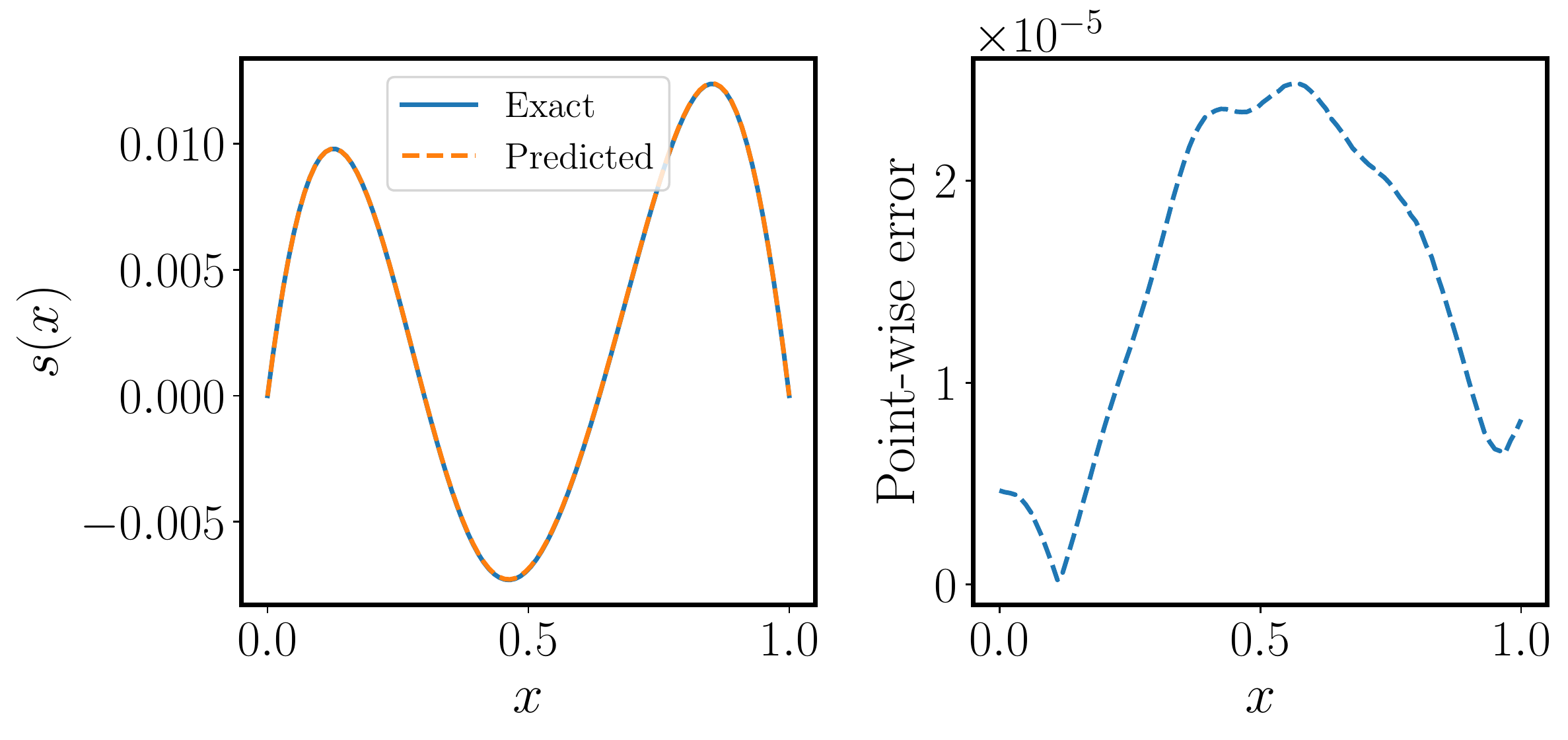}
     \end{subfigure}
          \begin{subfigure}[b]{0.45\textwidth}
         \centering
    \includegraphics[width=1.0\textwidth]{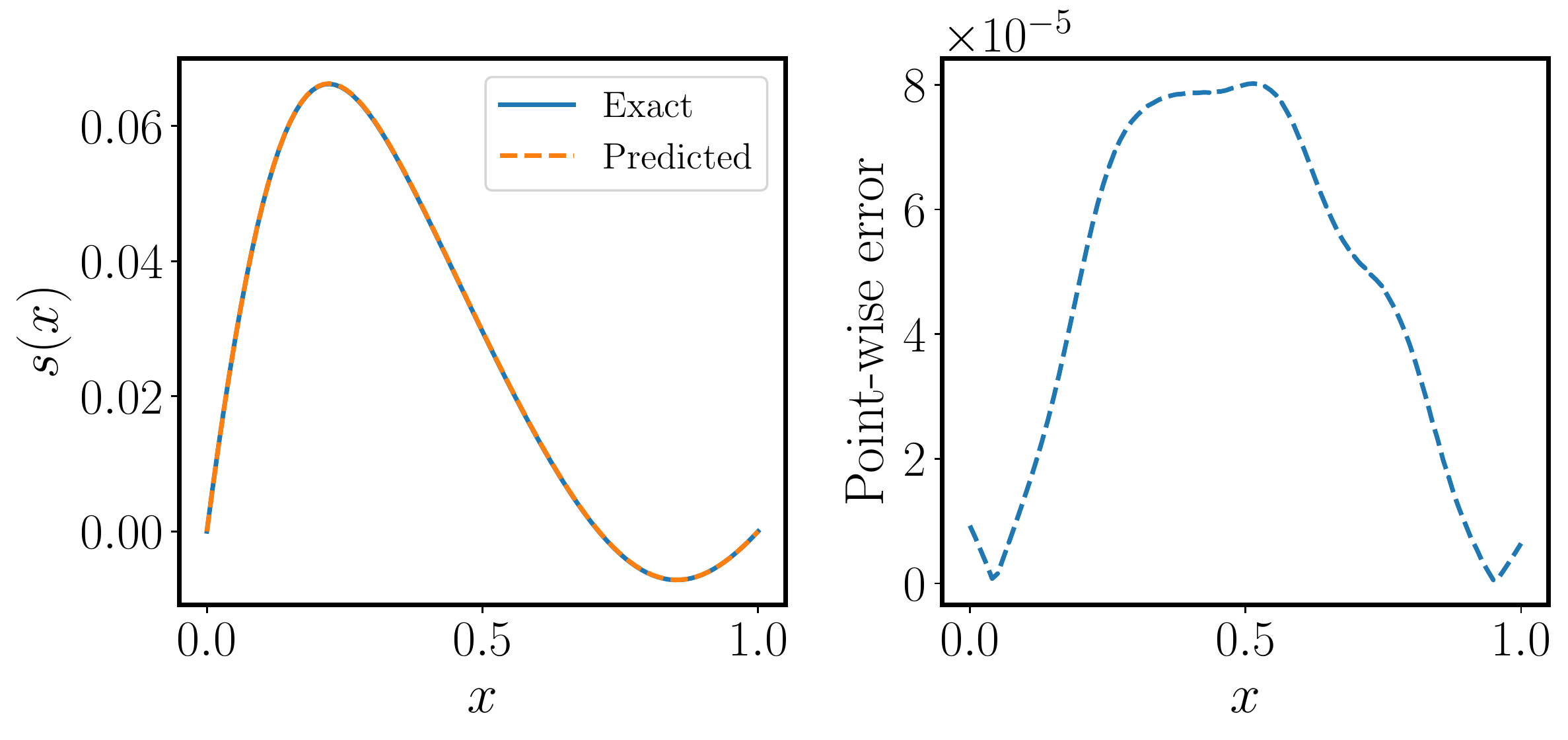}
     \end{subfigure}
      \caption{{\em 1D Poisson equation:} Predicted solutions of a trained physics-informed DeepONet for four different examples in the test data-set.}
        \label{fig: Poisson_pred}
\end{figure}

\clearpage
\subsection{2D Heat transfer}

\begin{figure}[ht]
     \centering
     \begin{subfigure}[b]{0.8\textwidth}
         \centering
    \includegraphics[width=1.0\textwidth]{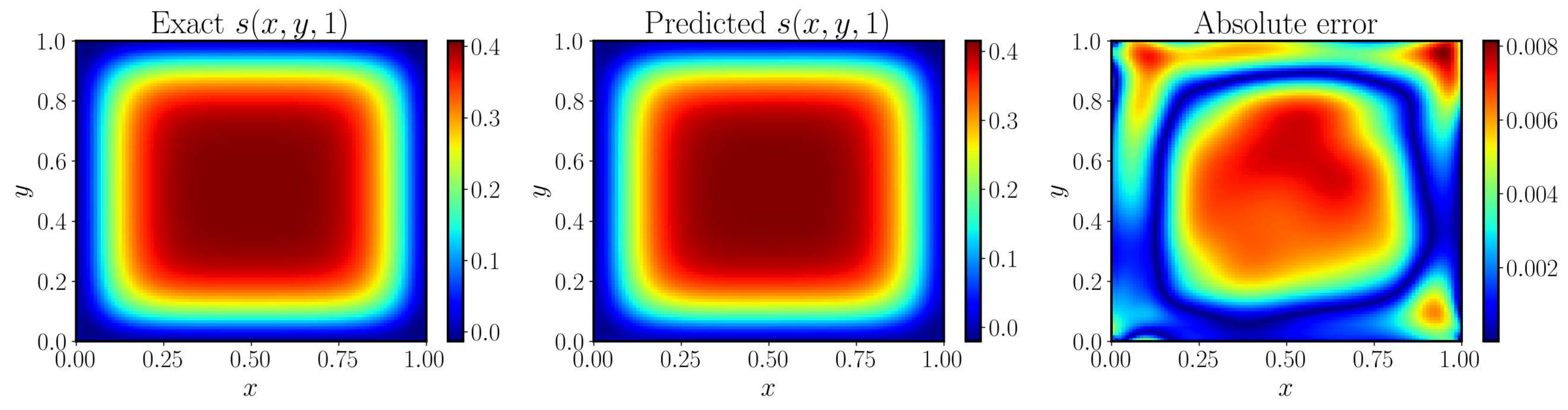}
     \end{subfigure}
          \begin{subfigure}[b]{0.8\textwidth}
         \centering
    \includegraphics[width=1.0\textwidth]{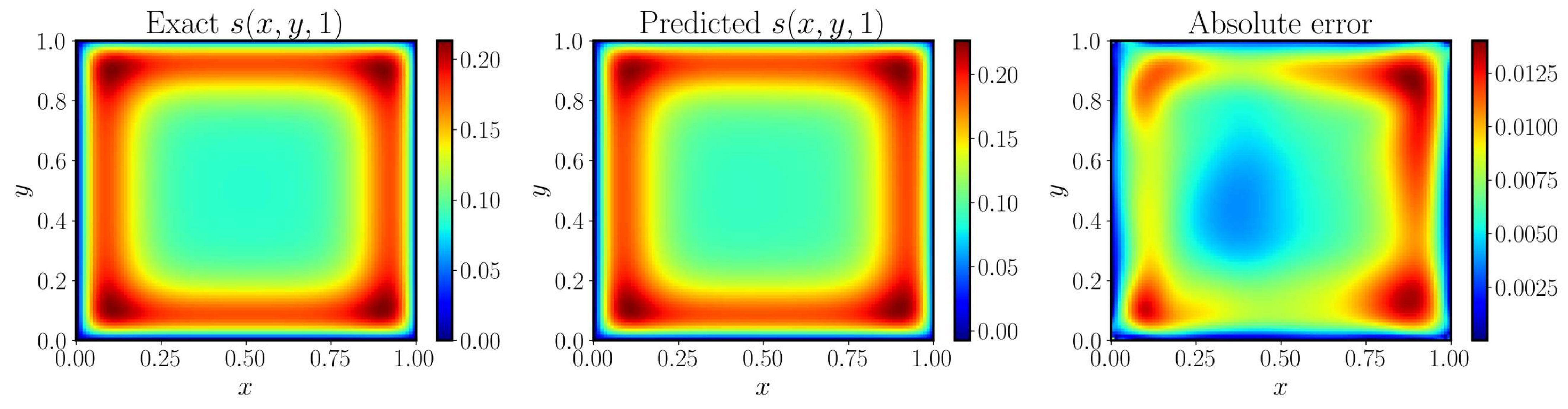}
     \end{subfigure}
          \begin{subfigure}[b]{0.8\textwidth}
         \centering
    \includegraphics[width=1.0\textwidth]{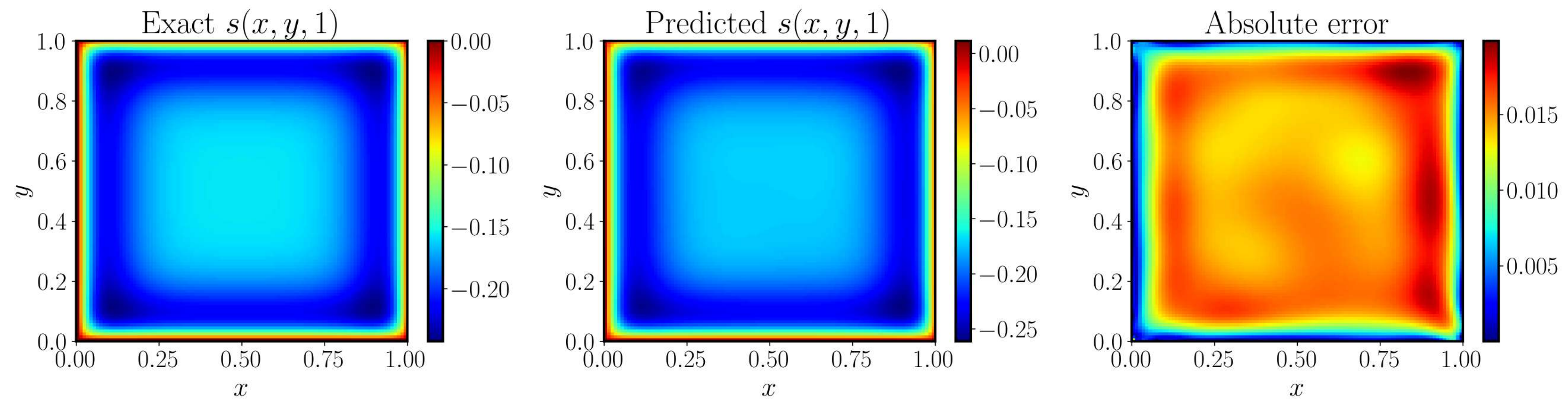}
     \end{subfigure}
       \caption{{\em 2D Heat equation:} Comparison of the predicted and exact solutions corresponding to the  temporal snapshots at $t=1$, for  three different examples in the test data-set.}
        \label{fig: heat_examples}
\end{figure}

\clearpage
\subsection{Stokes flow}

\begin{figure}[ht]
    \centering
    \includegraphics[width=0.8\textwidth]{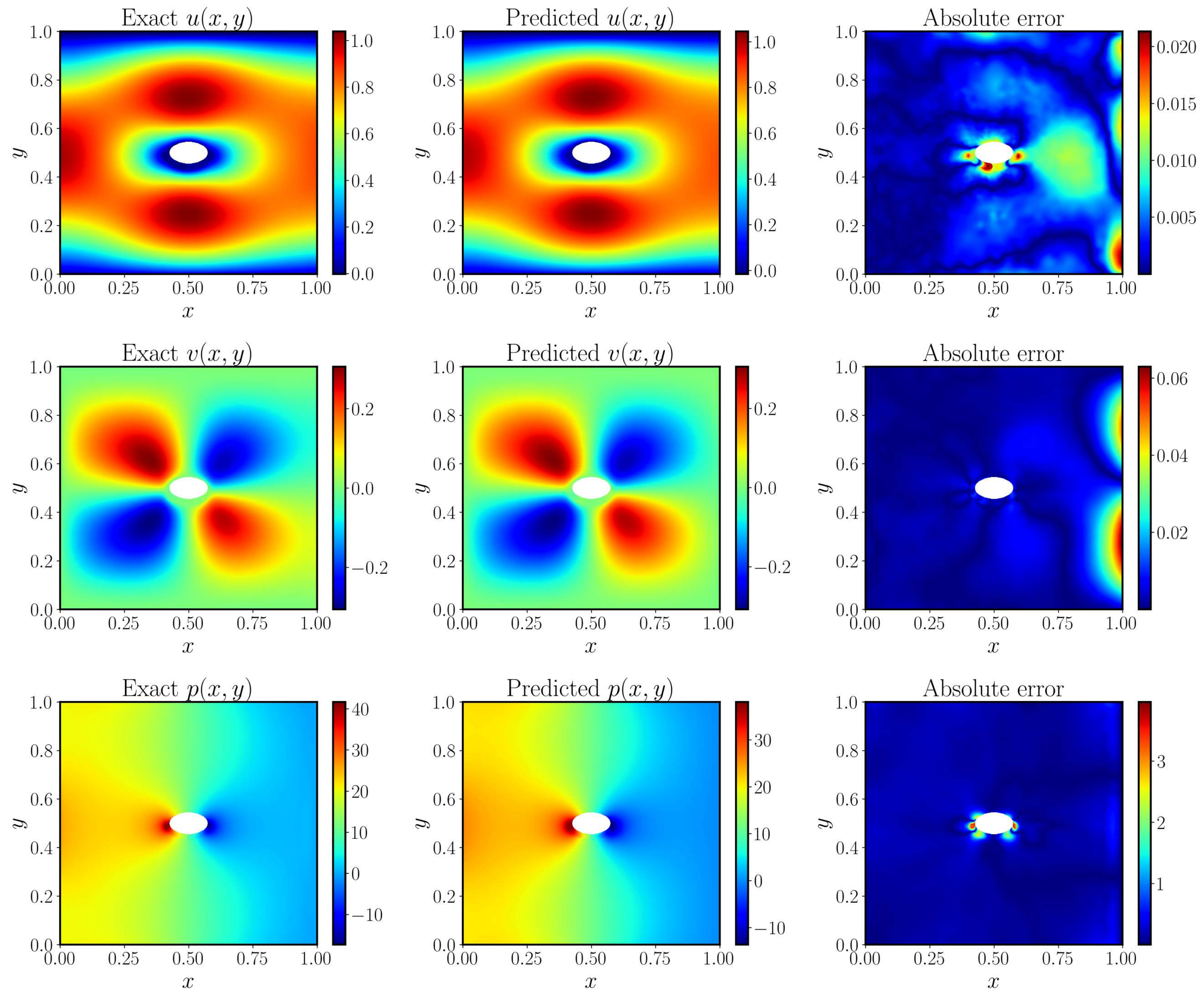}
    \caption{{\em Stokes flow:} Predicted solution of a  trained physics-informed DeepONet for one example in the test data-set.}
    \label{fig: Stokes_PI_deeponet_modified_deeponet_global_NTK_weights_pred_1}
\end{figure}

\begin{figure}[ht]
    \centering
    \includegraphics[width=0.8\textwidth]{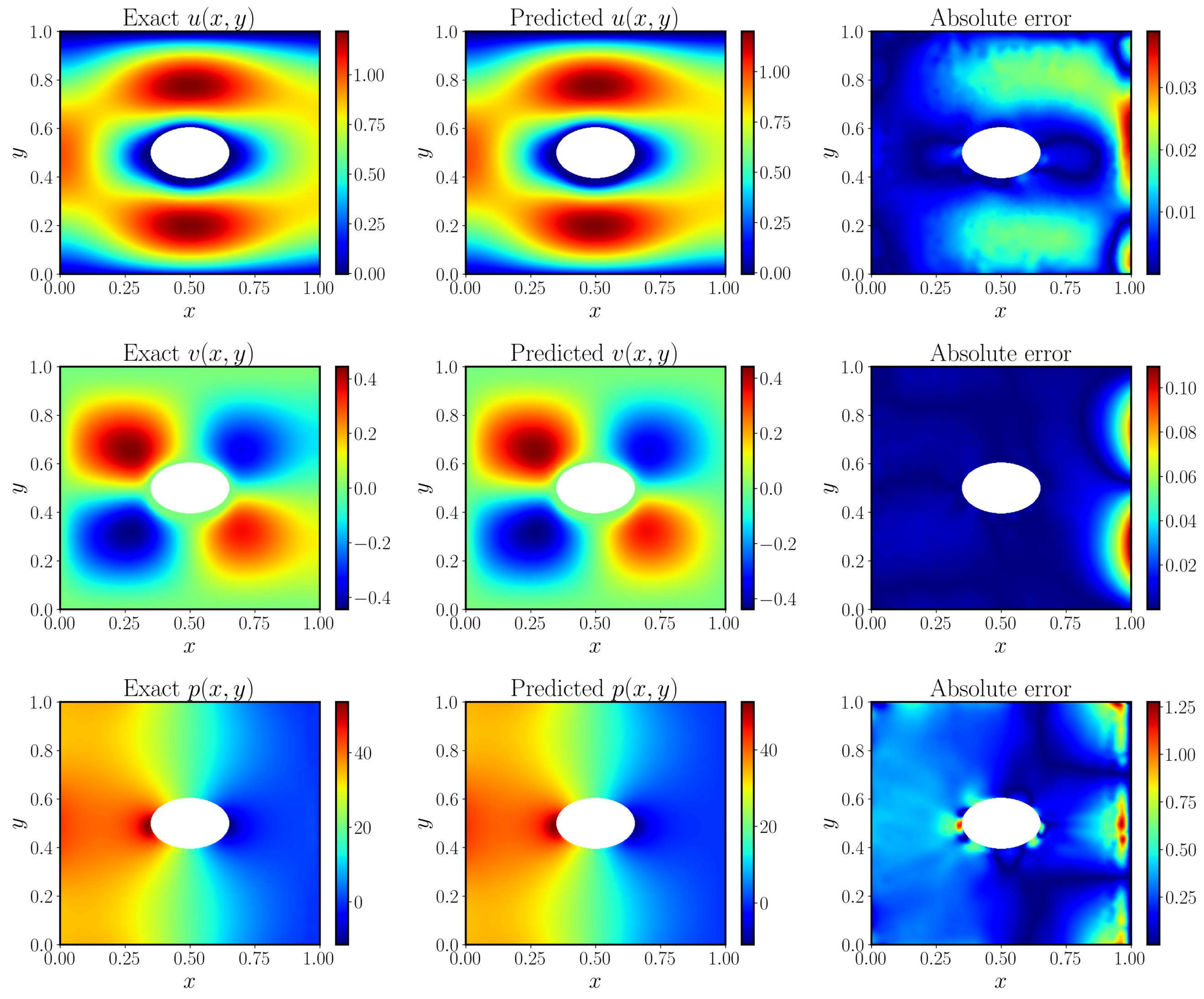}
    \caption{{\em Stokes flow:} Predicted solution of a  trained physics-informed DeepONet for one example in the test data-set.}
    \label{fig: Stokes_PI_deeponet_modified_deeponet_global_NTK_weights_pred_2}
\end{figure}

\begin{figure}[ht]
    \centering
    \includegraphics[width=0.8\textwidth]{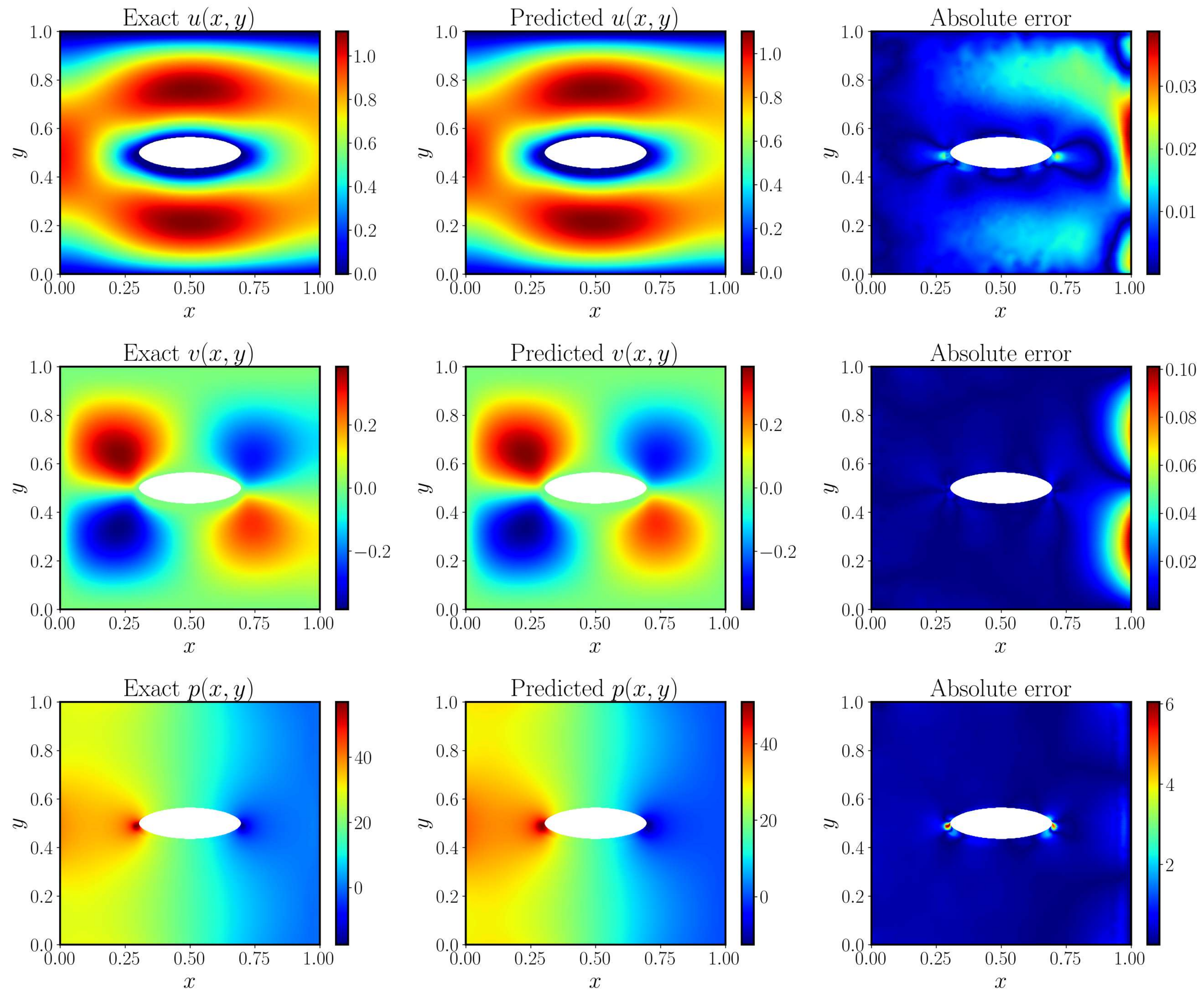}
    \caption{{\em Stokes flow:} Predicted solution of a  trained physics-informed DeepONet for one example in the test data-set.}
    \label{fig: Stokes_PI_deeponet_modified_deeponet_global_NTK_weights_pred_14}
\end{figure}

\end{document}